\documentclass[runningheads]{llncs}
\usepackage[T1]{fontenc}
\usepackage{hyperref}
\usepackage{url}

\usepackage[dvipsnames]{xcolor}
\usepackage{graphicx}
\usepackage{amsmath}
\usepackage{pifont}
\usepackage{amssymb}
\usepackage{booktabs}
\usepackage{subcaption}
\usepackage{verbatim}
\usepackage{physics}
\usepackage{multirow}
\usepackage{wrapfig}
\usepackage{float}

\newcommand{\review}[1]{{\textcolor{black}{#1}}}

\begin{document}
%
\title{Clustering-based Image-Text Graph Matching for Domain Generalization}

%
%
\author{Nokyung Park\inst{1} \and
Daewon Chae\inst{1} \and
Jeongyong Shim\inst{3} \and \\
Sangpil Kim\inst{2} \and
Eun-Sol Kim\inst{4, *} \and
Jinkyu Kim\inst{1, *}
}
\footnotetext{Co-correspondence to: Jinkyu Kim <jinkyukim@korea.ac.kr> and Eun-Sol Kim <eunsolkim@hanyang.ac.kr>.}
\authorrunning{N. Park et al.}
%
\institute{Department of Computer Science and Engineering, Korea University$^{1}$\\ 
Department of Artificial Intelligence, Korea University$^{2}$\\
Department of Artificial Intelligence Application, Hanyang University$^{3}$\\ 
Department of Computer Science, Hanyang University$^{4}$
}
%
\maketitle              
\begin{abstract}
   Learning domain-invariant visual representations is important to train a model that can generalize well to unseen target task domains. Recent works demonstrate that text descriptions contain high-level class-discriminative information and such auxiliary semantic cues can be used as effective pivot embedding for domain generalization problems. However, they use pivot embedding in a global manner (i.e., aligning an image embedding with sentence-level text embedding), which does not fully utilize the semantic cues of given text description. In this work, we advocate for the use of local alignment between image regions and corresponding textual descriptions to get domain-invariant features. To this end, we first represent image and text inputs as graphs. We then cluster nodes within these graphs and match the graph-based image node features to the nodes of textual graphs. This matching process is conducted both globally and locally, tightly aligning visual and textual semantic sub-structures. We experiment with large-scale public datasets, such as CUB-DG and DomainBed, and our model achieves matched or better state-of-the-art performance on these datasets. 
The code is available at: \href{https://github.com/noparkee/Graph-Clustering-based-DG}{https://github.com/noparkee/Graph-Clustering-based-DG}

\keywords{Domain Generalization  \and Multimodal Learning}
\end{abstract}
\section{Introduction}
\begin{figure}
    \centering
    \includegraphics[width=0.9\textwidth]
    {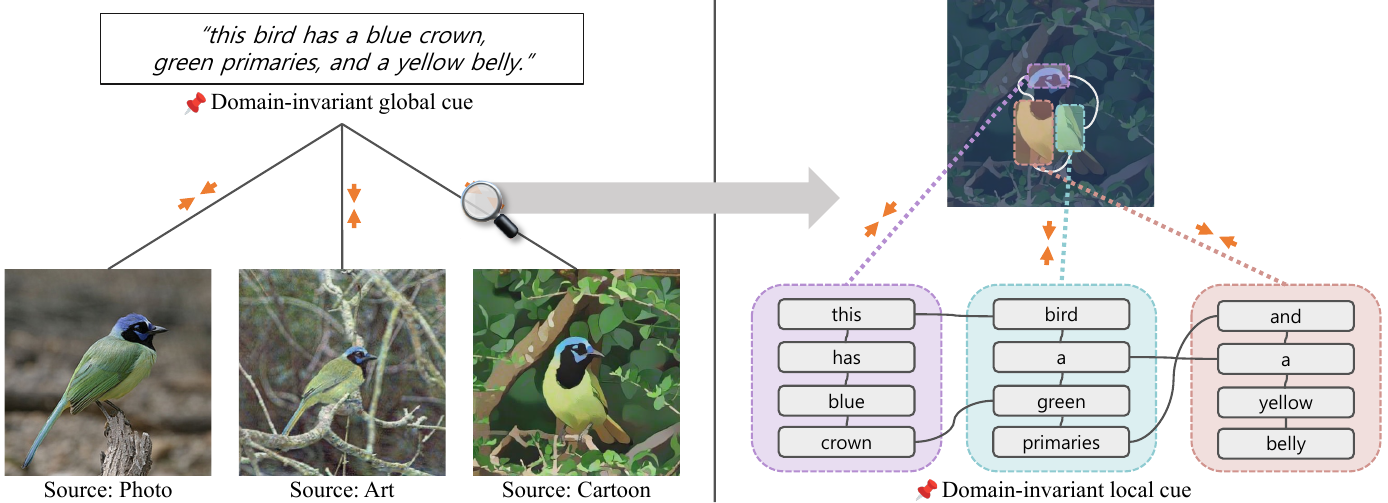}
    \vspace{-.5em}
    \caption{Our model learns domain-invariant visual representations by matching images and text descriptions at both global and local levels. Images and texts are represented as clustering-based graphs, encouraging the model to learn domain-invariant local semantic cues (e.g., ``a yellow belly'' and ``green primaries'').} 
    \label{fig:teaser}
    \vspace{-1.5em}
\end{figure}
%
How can humans effectively comprehend visual concepts despite variations in backgrounds, textures, and artistic styles? If it is impossible to collect sufficient examples of various combinations of domains, can current machine learning methods found on the i.i.d. assumption achieve robust generalization performance across domains? In this paper, we consider the domain generalization problem on image datasets and introduce a novel clustering-based image-text graph matching to tackle the problem.

Domain generalization aims to improve a model's generalization ability for unseen task domains. Previous research has explored various approaches to address this challenge, including minimizing domain discrepancies in the visual feature space~\cite{coral,selfreg}, augmenting data to cover various domains~\cite{fourier,neophile}, and utilizing ensemble learning~\cite{swad,cross_distillation}. Notably, recent work such as GVRT~\cite{gvrt} suggests leveraging natural language descriptions (e.g., ``this bird has a blue crown, green primaries, and a yellow belly'')  to infuse visual encoders with domain-invariant semantic cues, i.e., a visual encoder is optimized to produce an embedding that aligns well with the corresponding text embedding. While promising, optimizing a model with such global alignment often leads to suboptimal results, as these models may lack diverse attribute focus and occasionally attend to irrelevant regions for the class (e.g., see Figure~\ref{fig:gvrt_gradcam}).

To address these limitations, as shown in Figure~\ref{fig:teaser}, we focus on local matching, wherein image regions are matched with corresponding textual descriptions (e.g., an image region of blue crown and ``blue crown'' in a sentence). This approach involves representing the text descriptions with graphs and aligning the embedding of images and text by matching the graphs. 

As shown in Figure~\ref{fig:architecture}, the suggested method consists of three parts: (i) a graph-based visual encoder, (ii) a graph-based textual encoder, and (iii) a graph-based alignment.
Based on the graph-based representations, we aim to learn the domain-invariant features by grounding the graph-based image features into textual graphs, as the textual graphs contain explicitly verbalized knowledge from humans' typical reasoning.
To solve the language grounding with structural information, we suggest a new method that clusters the graph node features then matches those clusters.
By matching the multimodal graphs while clustering each node's features,  our suggested method can get robust domain-invariant features representing multilevel semantic alignment.

Experimental results with two popular benchmark datasets, CUB-DG~\cite{gvrt} and DomainBed~\cite{gulrajani2020search}, show the pivotal role of multimodal structural representations.
Quantitatively, our suggested method achieves a new state-of-the-art performance, especially by increasing generalization ability on the most difficult domain \textit{paint} of CUB-DG dataset.
With robust qualitative visualization results, we argue that our model learns domain-invariant features across various feature resolutions by locally and globally aligning with textual graphs.

Our contributions can be summarized as follows. (1) We propose the first approach using graph representations for both image and text inputs for the DG problem. (2) We suggest a novel method that clusters and matches node features to align two multimodal graphs. (3) We achieve a new state-of-the-art DG performance on the CUB-DG dataset and DomainBed benchmark.
\vspace{-.3em}

\section{Related Work}
\vspace{-0.3em}
\subsubsection{Domain Generalization.} Domain generalization aims to enhance a model's ability to generalize to unseen target domains with different data distributions compared to the source domains. The main idea of domain generalization is to learn domain-invariant features from multiple source domains. Various methods have been proposed to resolve this problem (i) by reducing domain discrepancies in the feature space~\cite{coral,selfreg}, (ii) by implementing data augmentation~\cite{fourier,neophile}, and (iii) by utilizing ensemble learning~\cite{swad}. (iv) Other studies have proposed using auxiliary semantic cues to facilitate learning domain-invariant features~\cite{decaug,lens,miro}.

Recently, GVRT~\cite{gvrt} successfully leverages textual descriptions for models to learn domain-invariant visual representations by aligning them with verbalized (domain-invariant and class-discriminative) knowledge from humans' typical reasoning (e.g., given a text ``this bird is black with an orange spot on its wing''). 
GVRT improves the model's generalization power by leveraging visual and textual inputs together and simply matching global representations. However, our focus extends beyond this, emphasizing the alignment of locally-aware high-order semantic relations via graph structures.

\vspace{-1em}
\subsubsection{Graph Neural Network.} 
Along with the huge success of neural networks in computer vision and natural language processing domains, new methodologies to deal with irregular structural inputs have been recently suggested. Various graph-based neural network algorithms are recommended for learning representations from structural inputs like molecular graphs, social networks, and meshes. According to the ways of representing graph data, attention-based methods (e.g., MoNet~\cite{monet}), convolution-based methods (e.g., GCN~\cite{gcn}), and message-passing methods (e.g., MPNN~\cite{mpnn}) can be applied to graph representation learning. Those graph neural network methods have achieved great performance on graph-related tasks, such as node classification~\cite{nodeclassification}, link prediction~\cite{linkprediction}, and graph classification problems~\cite{graphclassification}, by leveraging the non-euclidean data manifolds to get informative representations. Recently, the applications of graph neural networks have been extended to image and text domains~\cite{imagetextgraph}. By representing the image and text inputs as graphs, it becomes possible to consider the irregular and high-order correlations between tokens. In this paper, we suggest representing the multimodal inputs as graphs and matching the semantic correspondences between the multimodal inputs using graph neural networks to get the domain-invariant features.

\section{Method}
\begin{figure}[t]
    \includegraphics[width=\textwidth]{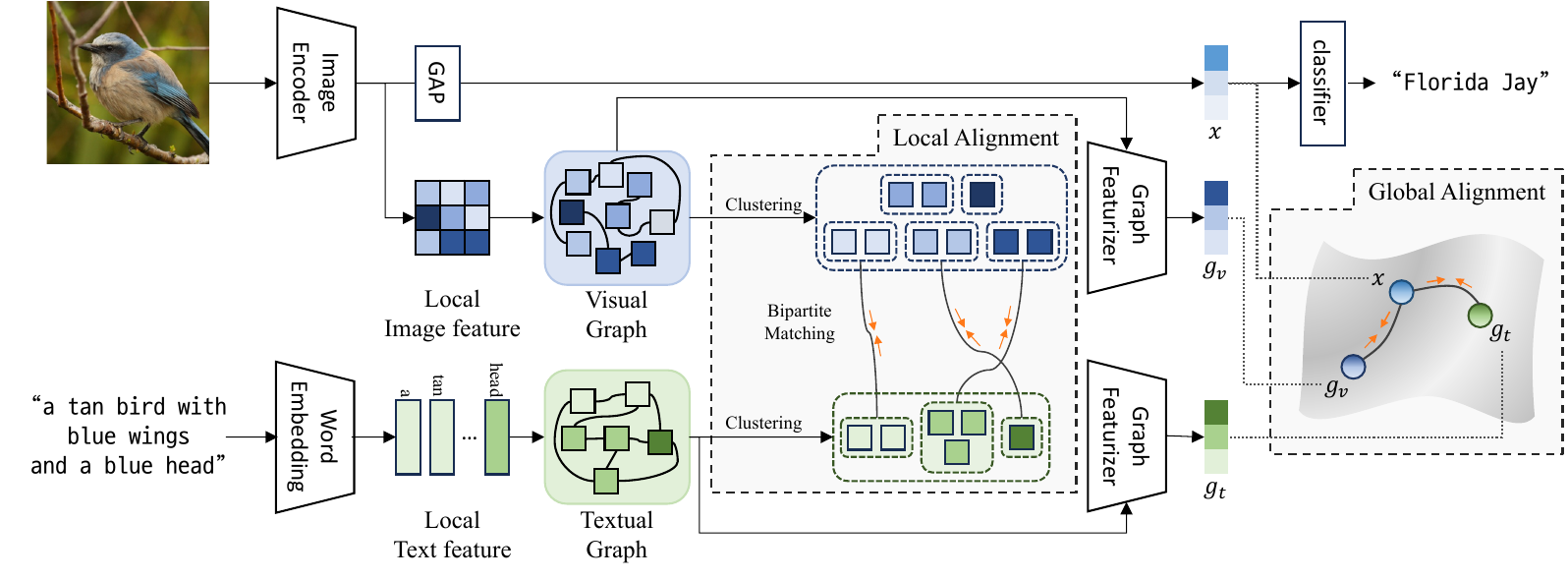}
    \vspace{-2.5em}
    \caption{An overview of our proposed method. We introduce multimodal graphs (visual and textual) that align with each other locally and globally,  yielding domain-invariant visual features that are well-aligned with humans' explicitly verbalized knowledge.}
    \label{fig:architecture}
    \vspace{-1em}
\end{figure}
Given a distribution over multiple (or single) source domains $\{\mathcal{S}_1, \mathcal{S}_2, \dots\}\in\mathcal{S}$, the domain generalization (DG) problem considers the following classical stochastic optimization, in which we minimize the data-dependent generalization upper bound of the expected task loss~\cite{sinha2017certifying}:
\begin{equation}
    \underset{\theta}{\mathrm{minimize}}~\underset{\mathcal{T: D(\mathcal{S}, \mathcal{T})\leq\rho}}{\mathrm{sup}}~\mathbb{E}_{\mathcal{T}}\big[ \mathcal{L}(\theta; \mathcal{S}) \big]
\end{equation}

where we consider unseen target domains $\mathcal{T}=\{\mathcal{T}_1, \mathcal{T}_2, \dots\}$ and the discrepancy between $\mathcal{S}$ and $\mathcal{T}$ is bounded by an arbitrary bound $\rho$, i.e. $D(\mathcal{S},\mathcal{T})\leq\rho$. We consider image classification scenarios and define the task-specific loss $\mathcal{L}$ function by the cross-entropy loss. Extracting domain-invariant representations from an image is key for training a model to generalize well to unseen domains.

Inspired by recent work by~\cite{gvrt}, we also want to improve the model's generalization power by leveraging visual and textual inputs together. Our model learns to extract (domain-invariant) visual representations that are well-aligned with explicitly verbalized knowledge from humans' typical reasoning. Unlike from~\cite{gvrt} in that we focus more on aligning locally-aware high-order semantic relations via graphs instead of simply aligning global representations.

Figure~\ref{fig:architecture} illustrates the architecture of our model, which is composed of three primary components: (i) a Graph-based Visual Encoder, (ii) a Graph-based Textual Encoder, and (iii) Graph-based Alignment for Learning Domain-Invariant Features. In (i), local latent representations (from a backbone network) are represented as a graph structure. Each local latent vector becomes a node, creating edges based on pairwise similarity in the embedding space (see Section~\ref{sec:visual_graph}). In (ii), we build a textual graph given a natural language description about a specific class (e.g., if the image corresponds to the \textit{Florida Jay} class, the description could be ``a tan bird with blue wings and a blue head.''). Each word embedding forms a node, creating edges based on embedding-level similarity (refer Section~\ref{sec:textual_graph}). In (iii), We regularize multi-modal encoders for locally aligned representations by minimizing graph-level distances between visual and textual data. Applying clustering-based graph matching makes our model generalizable
by learning human-compatible visual cues. (see Section~\ref{sec:alignment}).

\vspace{-0.5em}
\subsection{Graph-based Visual Encoder}
\label{sec:visual_graph}
\subsubsection{Global Visual Feature Extraction.}
Following standards in the domain generalization task, we use the pretrained ResNet50~\cite{resnet50} on ImageNet dataset~\cite{deng2009imagenet} as a backbone. Our backbone takes an image $\mathcal{I}$ as an input and produces a $d$-dimensional global visual representation $\mathbf{x}_g\in\mathbb{R}^{d}$. This global representation $\mathbf{x}_g$ is trained to predict its classification label ${\mathbf{y}}$ with a linear layer, yielding the per-class softmax probabilities $\hat{\mathbf{y}}$. Both the backbone and the classifier are trained by a classification loss $\mathcal{L}_c$ as follows:
\vspace{-0.3em}
\begin{equation}
    \mathcal{L}_c({\mathbf{y}}, \hat{{\mathbf{y}}})=-\sum_{i} y_i\text{log}(\hat{y}_i)
\vspace{-1em}
\end{equation}
where ${\bf{y}}\in\mathbb{R}^{|C|}$ represents the ground-truth one-hot vector, and $|C|$ denotes the size of the ground-truth class set. 
The model minimizes the loss function $\mathcal{L}_c$, but, unfortunately, this optimization often results in the model becoming semantically shallow.
Thus model would not generalize well in environments different from those in which they were trained. In our work, we aim to regularize our model to understand relations between visual cues and use those relations for the final verdict, thus making it more generalizable. We want to achieve such a regularization effect through utterances from human verbalized reasoning. 
%
\vspace{-1em}
\subsubsection{Locally-aware Visual Graph Construction.}
We first construct a graph with visual representations to achieve the abovementioned goal. Formally, given $M$ number of $d$-dimensional local visual representations 
$\mathbf{\textbf{x}}_l\in\{{\mathbf{x}}_{l,1}, {\mathbf{x}}_{l,2}, \dots, {\mathbf{x}}_{l,M}\}$ extracted from intermediate layers of the backbone (before global average pooling layer), we consider these representations as a set of unordered nodes. Note that each representation vector ${\mathbf{x}}_{l,i}\in\mathbb{R}^{d}$ for $i\in[1,M]$ corresponds to a certain grid over an input image $\mathcal{I}$.
Inspired by the recent work~\cite{visiongnn}, we construct a graph such that each node ${\mathbf{x}}_{l,i}$ has an edge with the other $K_v$ nearest neighbors (Visual Graph in Figure~\ref{fig:architecture}). We use the widely-used $L_2$ distance to measure pairwise node similarity. In summary, our visual graph $\mathcal{G}_v=(\mathcal{V}_v, \mathcal{E}_v)$, where $\mathcal{V}_v$ denotes the node set consisting of $M$ nodes (representing each visual feature of the local grid). The edge set $\mathcal{E}_v$ is comprised of the connections between nodes, with each node connected to its $K_v$ nearest neighbors. Detailed explanations are provided in the supplementary material.
%
\vspace{-1em}
\subsubsection{Graph-based Visual Representation.}
Given the visual graph $\mathcal{G}_v$, we further apply two layers of graph convolution network (GCN)~\cite{gcn}, each followed by a linear layer, BatchNorm~\cite{batchnorm} layer and ReLU~\cite{relu} activation. Subsequently, we employ dropout layer (only during training) and average readout operation to learn relational knowledge between local visual representations. Formally, we use a GCN-based function $f_{\text{GCN}}(\mathcal{G}_v)$ to obtain a final $d_g$-dimensional locally-aware visual graph representation ${\bf{g}}_v\in\mathbb{R}^{d_g}$, ${\bf{g}}_v = f_{\text{GCN}}(\mathcal{G}_v)$.
Note that, we also add an additional classifier that takes the $\textbf{g}_v$ as an input to create a graph that better captures the characteristics of the class. We provide detailed explanations in the supplementary material.
%
%
\vspace{-0.5em}
\subsection{Graph-based Textual Encoder}
\label{sec:textual_graph}
\subsubsection{Word-level Textual Graph Construction.} 
Our graph-based visual representation ${\bf{g}}_v$ encodes relational knowledge via a graph structure between representations of local visual features ${\bf{x}}_l$. We empirically observe that such graph-based representation's sole use is still insufficient for models to learn domain-invariant and human-compatible visual cues. Thus, to regularize our visual encoders to be aligned with human knowledge, we build a textual graph from a natural language description of each image, followed by aligning both visual and textual graphs. A sequence of $L$ (at maximum) words is first tokenized and encoded with a standard word-level (learnable) embedding layer, producing $d_t$-dimensional embedding vectors $\mathbf{t}\in\{\mathbf{t}_1, \mathbf{t}_2, \dots, \mathbf{t}_L\}$ where $\mathbf{t}_i\in\mathbb{R}^{d_t}$. Similar to our Visual Graph $\mathcal{G}_v$, we consider these word embeddings as an unordered set. We then construct a graph such that each node $\mathbf{t}_i$ has an edge with the other $K_t$ nearest neighbors (textual graph in Figure~\ref{fig:architecture}). We use $L_2$ distance to measure pairwise node similarity. Finally, we create a textual graph $\mathcal{G}_t=(\mathcal{V}_t, \mathcal{E}_t)$, where $\mathcal{V}_t$ denotes the node set comprising $L$ nodes, each representing a textual feature of word embedding, and $\mathcal{E}_t$ signifies the edge set.
%
\vspace{-1em}
\subsubsection{Graph-based Textual Representation.}
Given the textual graph $\mathcal{G}_t$, we apply the same architecture (but not shared) to obtain textual graph representation ${\bf{g}}_t\in\mathbb{R}^{d_g}$. I.e. we apply another GCN-based function $f_{\text{GCN}}(\mathcal{G}_t)$ to learn relational knowledge between word embeddings, ${\bf{g}}_t = f_{\text{GCN}}(\mathcal{G}_t)$. 
%
%
%
\vspace{-0.5em}
\subsection{Graph-based Alignment for Learning Domain-Invariant Features}
\label{sec:alignment}
We apply the following two graph-alignment approaches: (i) Graph-based Global Alignment and (ii) Local Alignment through Clustering-based Fine-grained Graph Matching, which comprises clustering and matching steps.
\vspace{-1em}
\subsubsection{Global Graph Alignment.}
We assume that text descriptions inherently contain class-discriminative semantic cues. Thus, our model can learn domain-invariant features with aid of textual information. A standard approach to aligning different representations is minimizing the Euclidean distance. We employ this alignment technique to graph features as follows: 
\vspace{-0.3em}
\begin{equation} \label{eq:coarse-grained}
    \begin{split}
    \mathcal{L}_\text{global} = ||f_{\text{proj},x}(\mathbf{x}_g) - f_{\text{proj},v}(\mathbf{g}_{v})||_2  + ||f_{\text{proj},x}(\mathbf{x}_g) - f_{\text{proj},t}(\mathbf{g}_{t})||_2
    \end{split}
\vspace{-1em}
\end{equation}

where we use a linear layer to project each feature (i.e. $\mathbf{x}_g$, $\mathbf{g}_v$, and $\mathbf{g}_t$) such that these three projected features are pulled together. Note that $f_{\text{proj},x}$, $f_{\text{proj},v}$, and $f_{\text{proj},t}$ represent a projection layer. Importantly, as we use only a force to pull latent representations together, the training dynamics may become unstable, causing a representation collapse. To avoid this, like the approach in~\cite{gvrt}, we add an auxiliary classifier which is trained with the standard cross-entropy loss takes $f_{\text{proj}, x}(\mathbf{x}_g)$ as an input to prevent a mode collapse, outputting the per-class softmax probability.
%
\vspace{-1em}
\subsubsection{Clustering Graph Nodes.}
In addition to global alignment, initially, we tried to match the nodes in the visual graph with textual graph to align the locally-aware semantic relations. However, simply aligning nodes from two different graphs may not work as these nodes have different representations (i.e. a visual feature of a local image region vs. a word-level representation). 
Therefore, we present clustering-based local graph matching, which applies a node clustering algorithm to ensure that the two graphs have the same level of semantic representation and then performs graph matching.
We define user-specified parameters $N_v(\leq M)$ and $N_t(\leq L)$ to the number of clusters for our visual and textual graphs, respectively. Note that we set $N_v \geq N_t$ since images may contain visual contents (e.g. backgrounds) that are not generally described in the text. 

Our approach to constructing a graph is based on measuring node similarity, which can result in a well-defined semantic structure in the graph. Therefore, we choose a modularity-based method for graph clustering that can reflect this semantic structure while remaining stable. Specifically, we use a deep learning-based modularity measurement method~\cite{dmon}. Our model first encodes the cluster assignment matrix using the features of the graph nodes. Then, we calculate the modularity using this matrix, which measures the quality of the clustering. We train the model to maximize the modularity while also constraining it with collapse regularization to prevent trivial solutions such as assigning all nodes to the same cluster.
We formulate it as follows: 
\vspace{-0.2em}
\begin{equation} 
\footnotesize{
    \mathcal{L}_{d} = -\frac{1}{2m}\Tr(\mathbf{C}^\mathrm{T}\mathbf{B}\mathbf{C})+ \frac{ \sqrt{k}}{n}\norm{\sum_{i}\mathbf{C}_{i}^\mathrm{T}}_{F}-1
    }
\vspace{-.2em}
\end{equation} 

where $\mathbf{C}$ is the cluster assignment matrix calculated with our graph feature, and $\mathbf{B}$ is the modularity matrix calculated with the adjacency matrix. ${m}$, ${n}$, and ${k}$ represent the number of edges, the number of nodes, and the number of clusters, respectively. The first term refers to modularity, which is measured using the assignment matrix and the modularity matrix, while the second term represents the collapse regularization term. So, our model can cluster semantically similar nodes together, allowing us to proceed with the matching process.

When applying node clustering to the visual graph $\mathcal{G}_v$, the cluster assignment matrix $\mathbf{C}$ has dimensions of $\mathbb{R}^{M \times N_v}$ (or $\mathbb{R}^{L \times N_t}$ when applied to the textual graph).
Each element in matrix $\mathbf{C}$ at the $i$-th row and $j$-th column represents the softmax probability that the $i$-th local feature (or the $i$-th node) belongs to the $j$-th cluster.
In this context, cluster feature is obtained through the following equation: $\mathcal{C}_v = f_{\text{proj},x}(\text{SeLU}({(\mathbf{C} / N_{v})}^{\mathrm{T}}  \textbf{x}_l)$), where $\text{SeLU}$~\cite{selu} serves as one of the activation functions.
Note that when dealing with the textual graph $\mathcal{G}_t$, $N_{v}$ and $\textbf{x}_l$ are replaced by $N_{t}$ and $\textbf{t}$, respectively.
%
\vspace{-1em}
\subsubsection{Clustering-based Graph Matching.}
Inspired by previous work~\cite{detr}, we use the set-based loss, i.e. the bipartite matching loss, between two disjoint sets of clusters: (i) a set of clusters $\mathcal{C}_v\in\{\mathcal{C}_v^1, \mathcal{C}_v^2, \dots,  \mathcal{C}_v^{N_v}\}$ of the visual graph $\mathcal{G}_v$ and (ii) a set of clusters $\mathcal{C}_t\in\{\mathcal{C}_t^1, \mathcal{C}_t^2, \dots, \mathcal{C}_t^{N_t}\}$ from the textual graph $\mathcal{G}_t$. We minimize the following pair-wise matching loss: 
%
\vspace{-0.5em}
\begin{equation}
   \mathcal{L}_{p} = \frac{1}{N_t}\sum_{i=1}^{N_t} ||\mathcal{C}_v^{{\mu}_i} - \mathcal{C}_t^{i} ||_2
\vspace{-.5em}
\end{equation}

where $\mu_i\in\{1,2,\dots,N_v\}$ is the cluster index of $\mathcal{C}_v$ which matches to $i$ in $\mathcal{C}_t$, producing the smallest total Euclidean distance by bipartite matching.

As the pair-wise matching loss pulls positive pairs together, negative pairs to add a repulsive force may need to prevent representation collapse. Thus, we also use a hinge loss based on $\mathcal{C}_v^i$ and $\mathcal{C'}_t^j$ (where $i\in[1,N_v]$ and $j\in[1,N_t]$), considering them as a negative pair if they are clusters for different input images. Thus, the matched distance $L_p$ should be smaller than any other pairs between $\mathcal{C}_v^{j}$ and $\mathcal{C'}_t^{i}$ (or $\mathcal{C'}_v^{j}$ and $\mathcal{C}_t^{i}$). We formulate it as a hinge loss as follows:
\vspace{-0.5em}

\begin{equation}
    \label{eq:hinge_loss}
    \begin{split}
        \mathcal{L}_h = \text{max}(0, \mathcal{L}_p - \text{MinDst}(\mathcal{C'}_v, \mathcal{C}_t) + \epsilon) + \text{max}(0, \mathcal{L}_p - \text{MinDst}(\mathcal{C}_v, \mathcal{C'}_t) + \epsilon)
    \end{split}
\end{equation}

where $\text{MinDst}(\mathcal{C}_v, \mathcal{C'}_t)$ and $\text{MinDst}(\mathcal{C'}_v, \mathcal{C}_t)$ represents the minimum pair-wise matching loss similar to $\mathcal{L}_p$, but is applied between two disjoint sets of clusters originating from different inputs within a mini-batch. We compute it across all sample pairs in a mini-batch and use the average as the final loss value:

\begin{equation}
    \mathcal{L}_\text{local} = \frac{1}{B}\sum_b (\lambda_d\mathcal{L}_d + \lambda_h\mathcal{L}_h + \lambda_{\text{aux}}\mathcal{L}_{\text{aux}})
\end{equation}
where we set the size of a mini-batch to $B$ and $\lambda_p$, $\lambda_h$, and $\lambda_d$ adjustable hyper-parameters that control the weight of each loss term. In our model, values of 1, 0.1, and 0.1 are used for $\lambda_d$, $\lambda_h$, and $\lambda_{\text{aux}}$, respectively.
Note that, similar to our global alignment module, we also add an auxiliary classifier that takes
the average-pooled representation of visual clusters matched with textual clusters, denoted as $\frac{1}{N_t}\sum_{i=1}^{N_t} \mathcal{C}_v^{{\mu}_i}$, as an input.
This classifier outputs the per-class softmax probability and is trained using the standard cross-entropy loss $\mathcal{L}_{\text{aux}}$.
%
\vspace{-1em}
\subsubsection{Loss Function.}
Ultimately, we train our model end-to-end by minimizing the following loss $L$:
\begin{equation} \label{eq:graph_level}
\vspace{-1em}
    \mathcal{L} = \mathcal{L}_{c} + \mathcal{L}_\text{global} + \mathcal{L}_\text{local}
\end{equation}
%
%

\vspace{-1.0em}
\section{Experiments}

\subsection{Setup}
\subsubsection{Implementation Details.}
Same as previous domain generalization approaches, we also use ResNet50~\cite{resnet50}, pre-trained on ImageNet~\cite{deng2009imagenet}, as our backbone, yielding a 2,048-dimensional visual representation from the last layer. Our model is trained end-to-end for 5,000 training steps using Adam optimizer with a learning rate of 5e-5. For training, we use standard image augmentation techniques such as random cropping, horizontal flipping, color jittering, grayscale conversion, and normalization. Our implementation is based on DomainBed~\cite{gulrajani2020search}, which is a unified domain generalization testbed, and our code will be publicly available upon publication. More details, including information that varies depending on the dataset, are available in supplementary material.

\vspace{-1em}
\subsubsection{Datasets.}
To demonstrate our model's effectiveness, we first use the CUB-DG dataset (for fine-grained image classification task), which is extended from the CUB dataset~\cite{WelinderEtal2010} for the domain generalization task.
This dataset contains 11,768 images for 200 classes of North American bird species. Each image has 10 text descriptions describing the content in detail, e.g., ``this bird is black with an orange spot on its wing.'' Each image is manipulated to create the following four domains: Photo, Cartoon, Art, and Painting.
We follow the common evaluation protocol and use the CUB-DG dataset's official split (the train and validation set has 5,994 samples, while the test set has 5,794 samples). 

Further, we also evaluate our model on DomainBed~\cite{gulrajani2020search}, which contains the following five multi-domain DG datasets: VLCS~\cite{fang2013unbiased}, PACS~\cite{Li2017dg}, OfficeHome~\cite{venkateswara2017deep}, TerraIncognita~\cite{beery2018recognition}, and DomainNet~\cite{leventidis2021domainnet}. Among these, we would emphasize that PACS~\cite{Li2017dg} dataset is useful for our experiments as (i) it provides a bigger domain shift than existing photo-only benchmarks, and (ii) it needs to exploit local information to learn discriminative subtle visual features. 
We follow the standard evaluation protocol used in \cite{gulrajani2020search}.
For datasets that do not provide text inputs, we use both (1) textual class definitions from the Oxford dictionary similar to GVRT~\cite{gvrt} and (2) descriptions generated by InstructBLIP~\cite{instructblip} with the prompt ``write a detailed description about the image.'' (refer supplementary material).
\vspace{-1em}


%
\subsection{Performance Comparison}
{
\setlength{\tabcolsep}{7pt}
    \begin{table}[t]
        \begin{center}
        \vspace{-1.0em}
        \caption{The out-of-distribution classification accuracies (in \%) on CUB-DG (top) and PACS (bottom) datasets based on the standard leave-one-out multi-source DG task setting.
        We compare ours with other existing DG approaches. (we provide full tables in supplementary material). \textit{Abbr.} I: Image, T: Text.}
        \vspace{.5em}
        \label{tab:performance_top5}
        \resizebox{1\linewidth}{!}{
        \begin{tabular}{@{}lcccccc@{}}   
        \toprule
        \multirow{2}{*}{Algorithms} & \multirow{2}{*}{Modality} &\multicolumn{4}{c}
        {Target Domain (Data: CUB-DG~\cite{gvrt})} & \multirow{2}{*}{Avg. $\uparrow$} \\\cmidrule{3-6}
        & & Photo & Cartoon & Art & Paint & \\\midrule
        
        MIRO~\cite{miro} & I & 68.2 & 59.1 & 46.5 & 38.2 & 53.0 \\
        SD~\cite{sd} & I & 71.3 & 62.2 & 50.8 & 34.8 & 54.7 \\
        CORAL~\cite{coral} & I & 72.2 & 63.5 & 50.3 & 35.8 & 55.4 \\
        \review{CCFP~\cite{CCFP}} & \review{I} & \review{70.0} & \review{61.5} & \review{52.1} & \review{\underline{40.4}} & \review{56.0} \\
        GVRT~\cite{gvrt} & I+T & \underline{74.6} & \underline{64.2} & \underline{52.2} & 37.0 & 57.0 \\
        Ours & I+T & \textbf{75.4} & \textbf{65.5} & \textbf{54.0} & \textbf{41.4} & \textbf{59.1} \\
        \bottomrule
        \toprule
        \multirow{2}{*}{Algorithms} & \multirow{2}{*}{Modality} &\multicolumn{4}{c}
        {Target Domain (Data: PACS~\cite{Li2017dg})} & \multirow{2}{*}{Avg. $\uparrow$} \\\cmidrule{3-6}
        & & Art Painting & Cartoon & Photo & Sketch & \\\midrule
        SelfReg~\cite{selfreg} & I & \underline{87.9 $\pm$ 1.0} & 79.4 $\pm$ 1.4  & 96.8 $\pm$ 0.7 & 78.3 $\pm$ 1.2 & 85.6\\
        CORAL~\cite{coral} & I & \textbf{88.3 $\pm$ 0.2} & 80.0 $\pm$ 0.5 & 97.5 $\pm$ 0.3 & 78.8 $\pm$ 1.3 & 86.2\\
        mDSDI~\cite{msdi} & I & 87.7 $\pm$ 0.4 & 80.4 $\pm$ 0.7 & \textbf{98.1 $\pm$ 0.3} & 78.4 $\pm$ 1.2 & 86.2\\
        SagNet~\cite{sagnet} & I & 87.4 $\pm$ 1.0 & 80.7 $\pm$ 0.6 & 97.1 $\pm$ 0.1 & 80.0 $\pm$ 0.4 & 86.3\\
        CCFP~\cite{CCFP} & I & 87.5 $\pm$ 0.1 & \underline{81.3 $\pm$ 0.3} & 96.4 $\pm$ 0.3 & \textbf{81.4 $\pm$ 0.8} & 86.6\\
        Ours & I+T & \underline{87.9 $\pm$ 0.7} & \textbf{81.4 $\pm$ 0.1} & \underline{98.0 $\pm$ 0.1} & \underline{80.5 $\pm$ 1.1} & \textbf{87.0}\\
        \bottomrule
        \end{tabular}
        }
        \end{center}
    \vspace{-2em}
    \end{table}
}

As shown in Table~\ref{tab:performance_top5}, we compared the out-of-distribution classification accuracies on the following two datasets: CUB-DG (top) and PACS (bottom) datasets. We compare ours with other existing state-of-the-art domain generalization approaches, SagNet~\cite{sagnet}, MIRO~\cite{miro}, SD~\cite{sd}, CORAL~\cite{coral}, GVRT~\cite{gvrt}, SelfReg~\cite{selfreg}, mDSDI~\cite{msdi}, \review{and CCFP~\cite{CCFP}}. Due to space constraints, we only report \review{top-6} results (we provide full tables in supplementary material).

As shown in Table~\ref{tab:performance_top5} (top), our proposed method clearly outperforms the other domain generalization techniques on the CUB-DG dataset in all target domains with a significant gain. In terms of the average image classification accuracy, ours shows 59.1\%, which is 2.1\% higher than GVRT~\cite{gvrt} (which uses the same image and text inputs) and 3.1\% higher than image-only approach, CCFP~\cite{CCFP}. Similar trends are also observed in our experiment on the large-scale PACS~\cite{Li2017dg} dataset. As shown in Table~\ref{tab:performance_top5} (bottom), our model also outperforms the other approaches, i.e., ours shows 87.0\% that is \review{0.4\% higher than CCFP~\cite{CCFP}-based SOTA approach} and 1.9\% higher than GVRT~\cite{gvrt}. These confirm that our graph-based approach is effective in aligning visual and textual encoders for fine-grained image classification tasks, improving the visual encoder's generalization power to unseen target domains. 

\subsection{Few-shot DG Performance Comparison}

\setlength{\tabcolsep}{3pt}
\renewcommand{\arraystretch}{1.0}
\begin{wraptable}{r}{.6\linewidth}
    \centering
    \vspace{-2.4em}
    \caption{Few-shot DG performance comparison.
    }
    \vspace{-0.7em}
    \resizebox{\linewidth}{!}{%
    \begin{tabular}{lc|lc}
        \toprule
        Algorithm (Data: PACS~\cite{Li2017dg}) & Avg. & Algorithm (Data: VLCS~\cite{fang2013unbiased}) & Avg. \\\midrule
        mDSDI~\cite{msdi} & 63.5 & mDSDI~\cite{msdi} & 68.5\\
        CORAL~\cite{coral} & 64.6 & GVRT~\cite{gvrt} & 69.4\\
        MIRO~\cite{miro} & 65.5 & MIRO~\cite{miro} & 69.6\\
        GVRT~\cite{gvrt} & 68.7 & CORAL~\cite{coral} & 71.1\\
        Ours & \textbf{70.7} & Ours & \textbf{71.4}\\
        \bottomrule
    \end{tabular}}
    \label{tab:few-shot}
    \vspace{-2.3em}
\end{wraptable}

Conventional DG approaches often assume that a sufficient number of images is available for all classes and domains enough to learn domain-invariant class-discriminative features. However, this may be practically challenging in real-world scenarios. We emphasize that our method, which leverages textual descriptions as pivotal information, can benefit learning domain-invariant features in the few-shot setting.
In Table~\ref{tab:few-shot}, our model significantly outperforms the other approaches on PACS~\cite{Li2017dg} and VLCS~\cite{fang2013unbiased} datasets. Note that we use randomly chosen five images (per class in each domain) as an input to train all models (i.e., 5-shot DG). Except for the number of training images, we generally follow the standard protocol of DomainBed for evaluation. 
%

%

\begin{figure}[t]
    \centering
    \includegraphics[width=.95\linewidth]
    {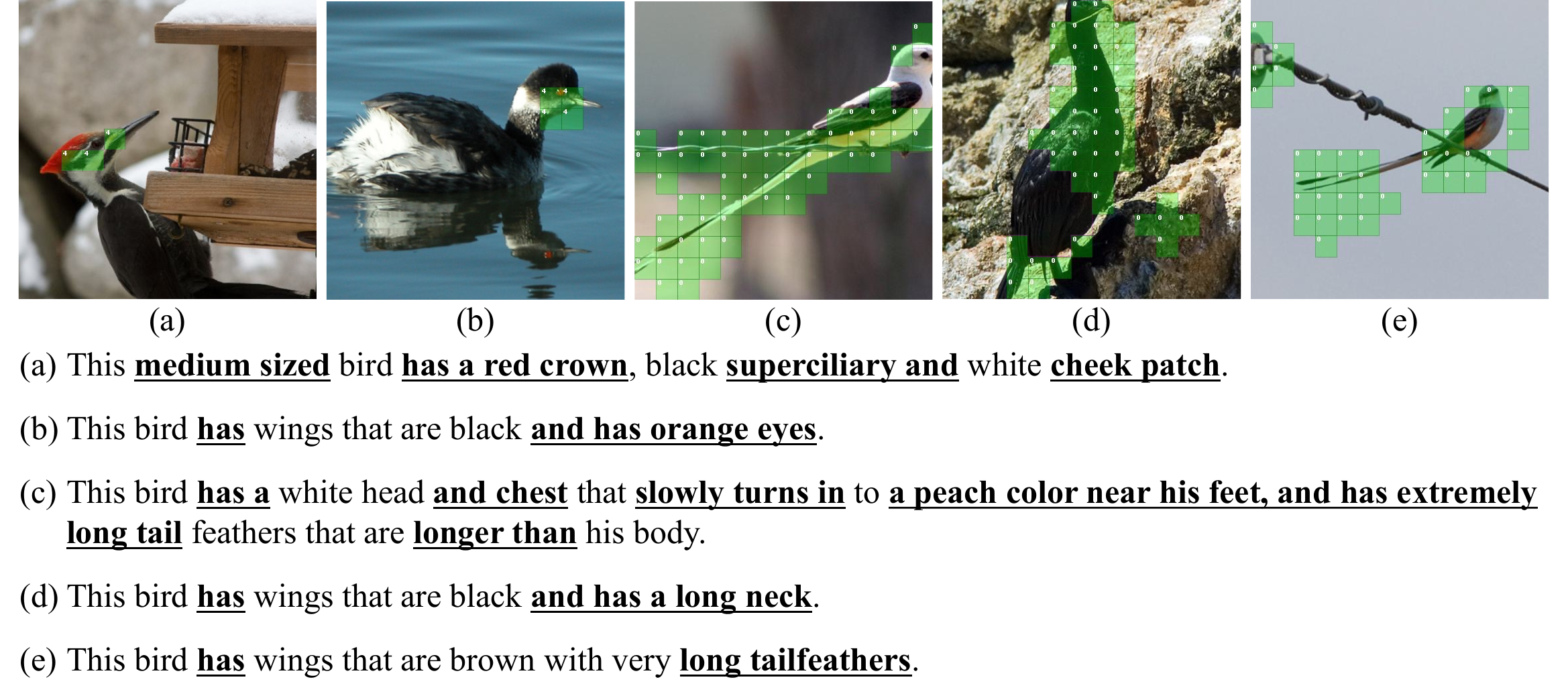}
    \vspace{-.5em}
    \caption{Examples of the matched image region (in visual graph clusters) and texts (in textual graph clusters).}
    \label{fig:graph_matching}
\end{figure}
\subsection{Analysis on CUB-DG Dataset}
\subsubsection{Analysis of Graph Clusters and Their Matchings.}
In Figure~\ref{fig:graph_matching}, we provide examples of a matched pair of image regions and a set of words. For example, in (a), a region around the bird's head is matched with a textual graph cluster that contains words including ``orange eyes''. We observe that our model reasonably matches image features with class-discriminative texts, e.g., red crown, cheek patch, extremely long tail, and long neck.
\subsubsection{t-SNE Analysis.} 
\begin{figure*}[t]
    \centering
    \vspace{.5em}
    \includegraphics[width=.95\linewidth]{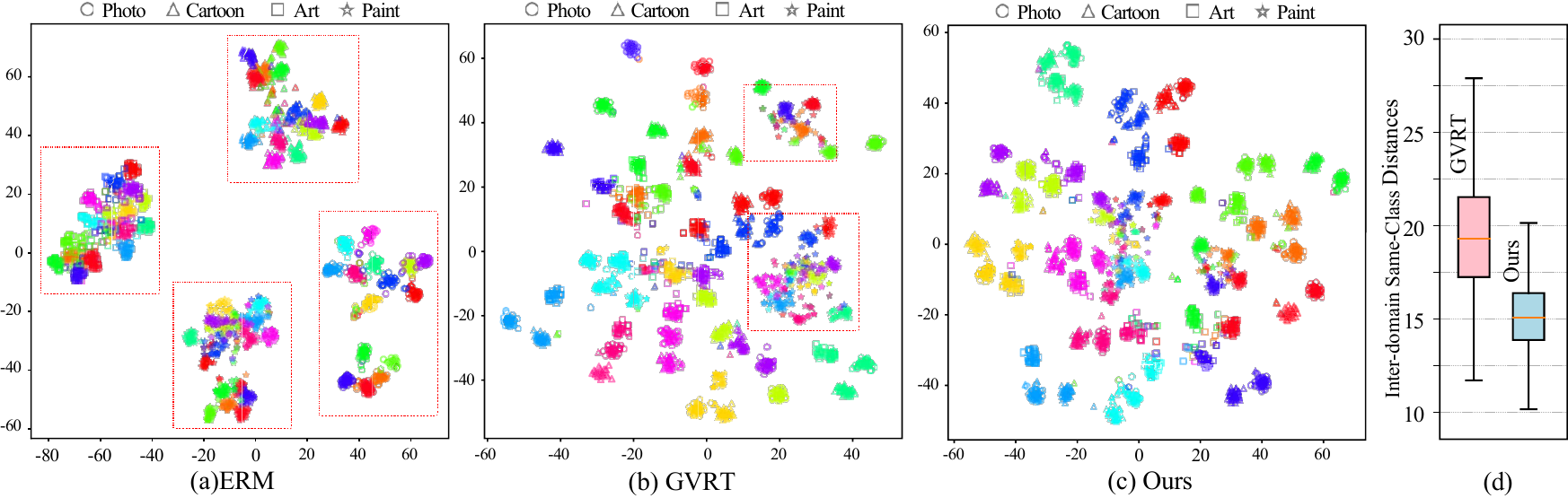}
    \vspace{-.5em}
    \caption{Visualizations by t-SNE for (a)ERM~\cite{erm}, (b)GVRT~\cite{gvrt}, and (c)Ours on CUB-DG. Points are color-coded differently by its class and has different shapes according to its domain. (d)We also compare inter-domain same-class distances.}
    \label{fig:tsne}
    \vspace{-1.0em}
\end{figure*}
As shown in Figure~\ref{fig:tsne}, we provide t-SNE~\cite{tsne} visualization of
\begin{wrapfigure}{r}{.45\linewidth}
    \centering
    \vspace{-2em}
    \includegraphics[width=\linewidth]    {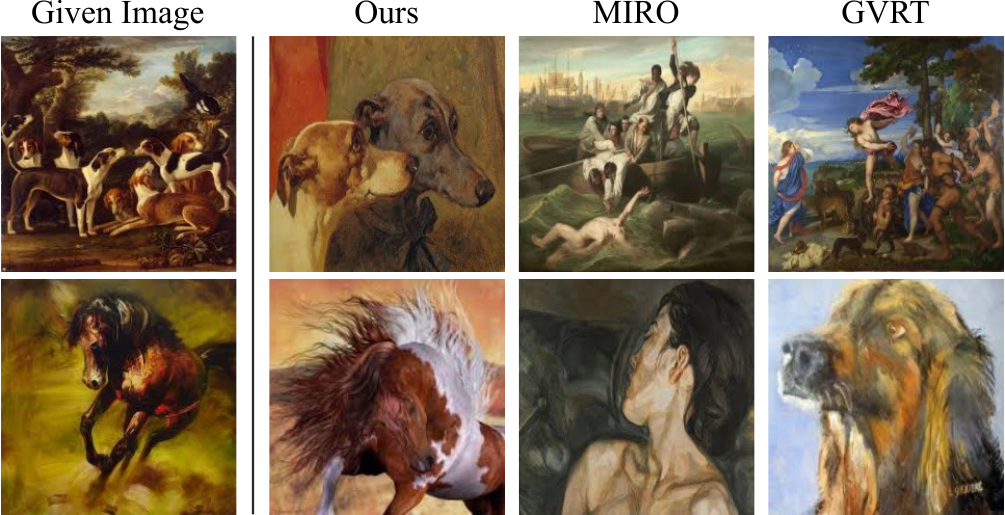}
    \caption{\review{Exemplars of the nearest examples from PACS dataset (in the unseen target domain) to the given image (e.g., ``dog'' and ``horse'').}}
    \label{fig:nearest_top1}
    \vspace{-3em}
\end{wrapfigure}
(a) ERM~\cite{erm}, (b) GVRT~\cite{gvrt}, and (c) Ours to visualize their embedding space on CUB-DG dataset. We use different marker styles (for target domains) and different colors (for classes).
An ideal model would show that visual features of the same class but different domains are gathered together. Ours clearly outperform ERM, which has scattered points per domain and better than GVRT in that features of the same class but different domains are more clustered (see red boxes). In Figure~\ref{fig:tsne} (d), we provide box plots for GVRT and ours, showing that our model produces lower same-class inter-domain distances than GVRT. Note that we provide detailed t-SNE visualizations in supplementary material.
\review{Figure~\ref{fig:nearest_top1} further shows the nearest examples in the unseen target domain to the given image (e.g., ``dog''). In contrast to MIRO and GVRT, which often provide examples of different classes (e.g., ``person''), ours consistently provide examples of the same class.
This is consistent with our t-SNE analysis. We provide more examples in in supplementary material.}

\vspace{-1.0em}
\subsubsection{GradCAM Visualization.}
\begin{wrapfigure}{t}{0.55\linewidth}
    \vspace{-2.2em}
    \begin{subfigure}{0.49\linewidth}
        \centering
        \includegraphics[width=1.0\linewidth]{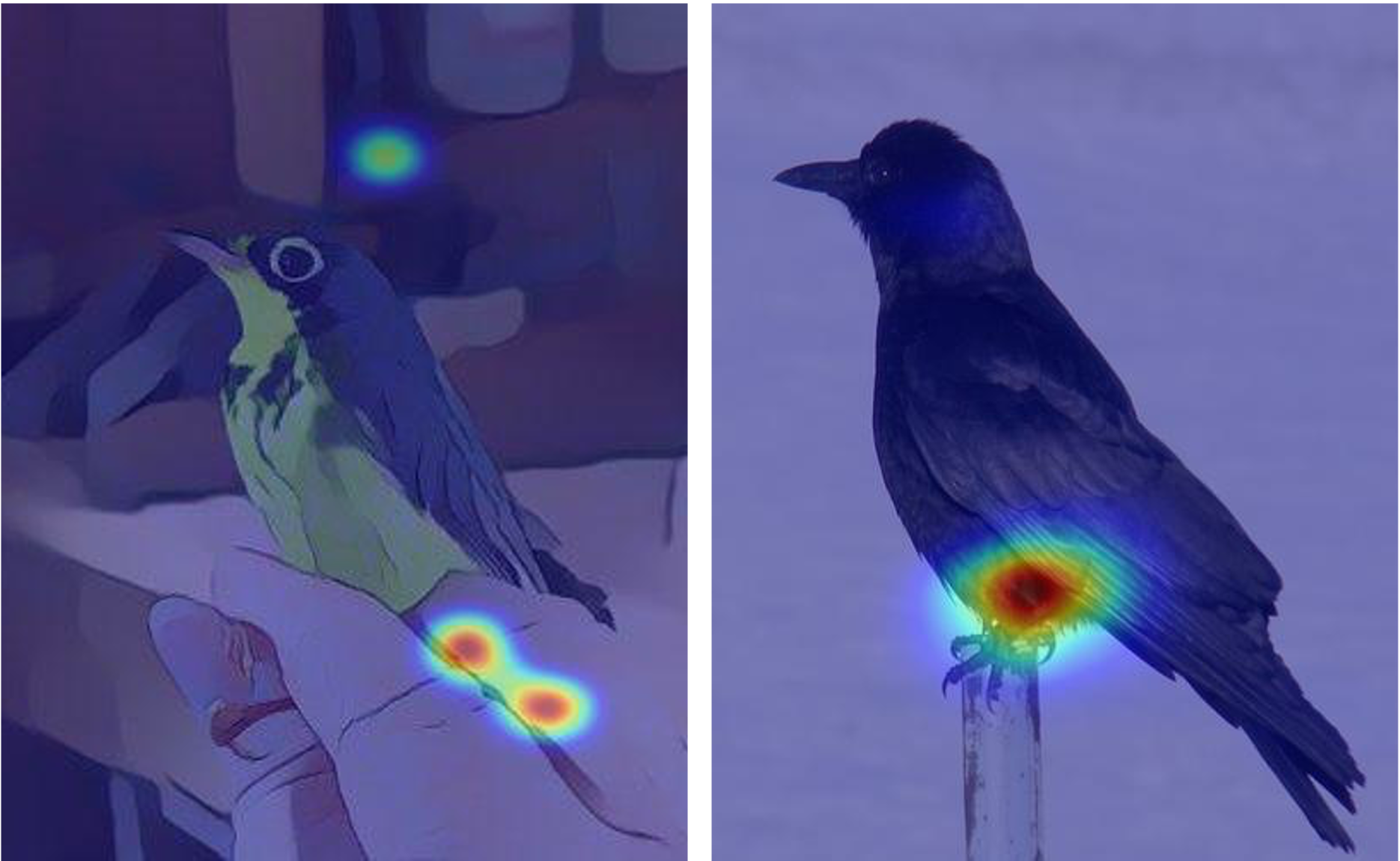}
        \caption{GVRT}
        \label{fig:gvrt_gradcam}
    \end{subfigure}
    \hfill
    \begin{subfigure}{0.49\linewidth}
        \centering
        \includegraphics[width=1.0\linewidth]{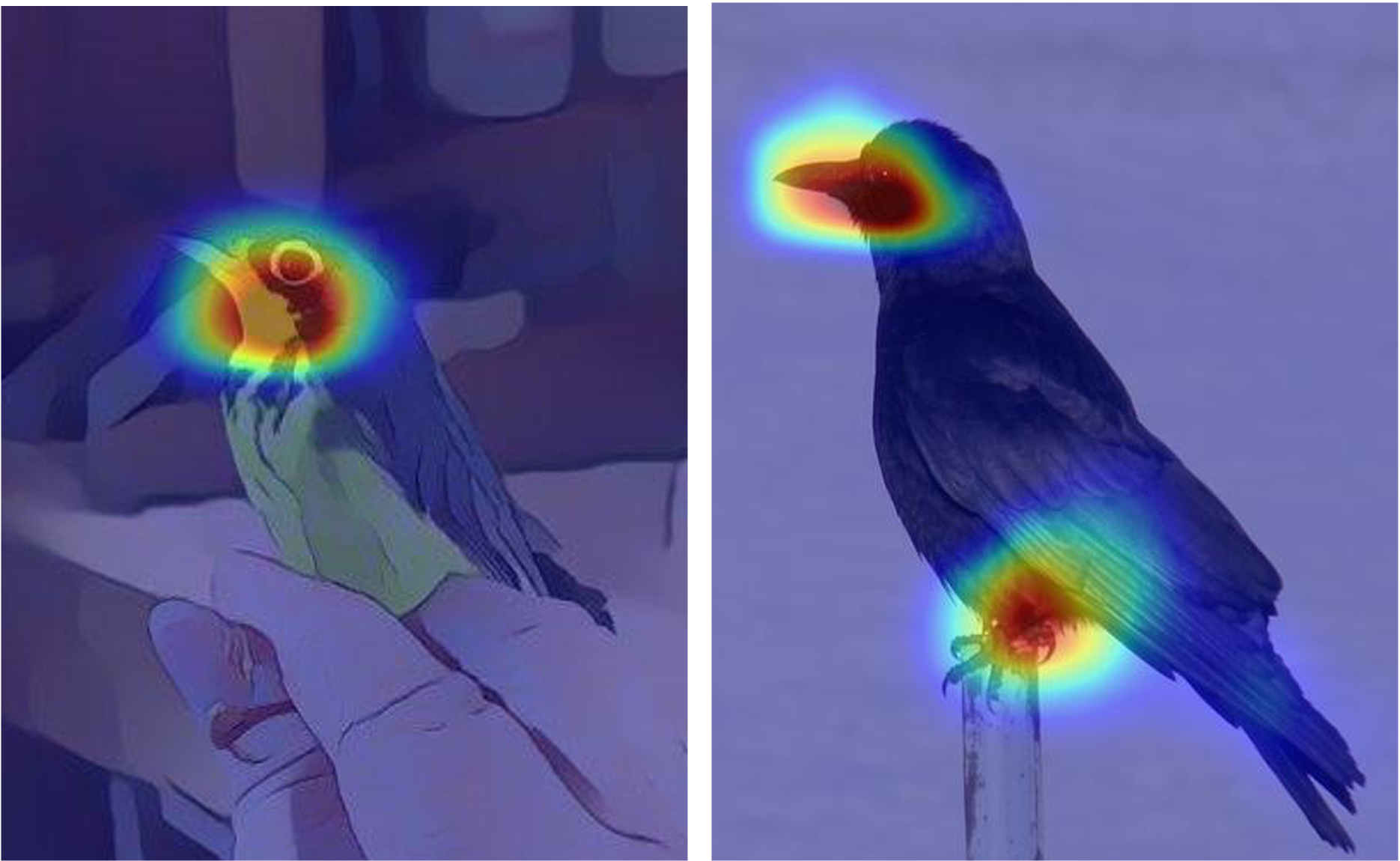}
        \caption{Ours}
        \label{fig:ours_gradcam}
    \end{subfigure}
    \vspace{-2.0em}
    \caption{GradCAM~\cite{grad_cam} visualizations to evaluate where the model sees.}
    \vspace{-2em}
  \label{fig:grad_cam}
\end{wrapfigure}
In Figure~\ref{fig:grad_cam}, we use GradCAM~\cite{grad_cam} to visualize image regions where the model focuses on for the final verdict. We observe that our model generally focuses on multiple class-discriminative features, giving the benefits of more robust and generalizable recognition performance. More specifically, in the first image of Figure~\ref{fig:ours_gradcam}, our model focuses on the head region, a relevant area for the classification task, in contrast to the GVRT model which attends to a region less relevant to class information (Figure~\ref{fig:gvrt_gradcam}). 
Furthermore, in the second image of the GVRT scenario, the model concentrates
solely on the belly region, while our model exhibits a more diverse focus on attributes, encompassing both the belly and beak.
Unlike GVRT, which focuses solely on global alignment, our model incorporates local alignment through graphs. This approach enables our model to capture diverse attributes, increasing its ability to generalize effectively.

%
\setlength{\tabcolsep}{7pt}
\renewcommand{\arraystretch}{1.0}
\begin{table}[t]
    \centering
    \caption{Ablation studies to compare variants of our model. Data: CUB-DG.}
    \label{tab:ablation-module}
    \resizebox{0.95\linewidth}{!}{%
    \begin{tabular}{cc|ccccccc}
        \toprule
        \multirow{2}{*}{\shortstack{Global \\ Alignment}} &
        \multirow{2}{*}{\shortstack{Local \\ Alignment}} &
        \multirow{2}{*}{\shortstack{Visual\\Graph}} &
        \multirow{2}{*}{\shortstack{Textual\\Graph}} &
        \multicolumn{4}{c}{Target Domain} &
        \multirow{2}{*}{Avg.} \\
        \cmidrule{5-8}
        & & & & Photo & Cartoon & Art & Paint &  \\\midrule

        - & - & \checkmark & \checkmark & 65.1 & 52.5 & 38.2 & 29.0 & 46.2 \\\midrule
        \checkmark & - & - & - & 69.5 & 57.1 & 44.2 & 30.2 & 50.2 \\
        \checkmark & - & \checkmark & - & 70.3 & 57.0 & 48.1 & 33.5 & 52.2 \\
        \checkmark & - & - & \checkmark & 75.0 & 64.4 & 53.0 & 34.7 & 56.8 \\
        \midrule
        \review{-} & \review{\checkmark} & \review{\checkmark} & \review{\checkmark} & \review{71.4} & \review{57.6} & \review{46.6} & \review{37.2} & \review{53.2} \\
        \midrule
        \checkmark & \checkmark & \checkmark & \checkmark & \textbf{75.4} & \textbf{65.5} & \textbf{54.0} & \textbf{41.4} & \textbf{59.1} \\
        \bottomrule
    \end{tabular}}
    
    \vspace{-1em}
\end{table}

{
\setlength{\tabcolsep}{1.9pt}
\renewcommand{\arraystretch}{1.1} 
    \begin{table}[t]
        \begin{center}
        \caption{The test accuracies (in \%) on the DomainBed datasets in the multi-source DG task setting. We compare ours with other existing DG approaches (we provide full tables in supplementary material). $\dagger$: utilizing texts from a dictionary, $\ddagger$: incorporating texts from InstructBLIP. \textit{Abbr.} I: Image, T: Text. 
        }
        \label{tab:performance_domainbed}
        \resizebox{0.95\linewidth}{!}{%
        \begin{tabular}{@{}lccccccc@{}}    \toprule
            \multirow{2}{*}{Algorithm} & \multirow{2}{*}{Modality} & \multicolumn{5}{c}{Dataset} & \multirow{2}{*}{Avg.} \\\cmidrule{3-7}
            & & VLCS & PACS & OfficeHome & TerraIncognita & DomainNet \\\midrule
            CORAL~\cite{coral} & I & 78.8 $\pm$ 0.6 & 86.2 $\pm$ 0.3 & 68.7 $\pm$ 0.3 & 47.6 $\pm$ 1.0 & 41.5 $\pm$ 0.1 & 64.6\\
            \review{CCFP~\cite{CCFP}}& \review{I} & \review{78.9 $\pm$ 0.3} & \review{86.6 $\pm$ 0.2} & \review{68.9 $\pm$ 0.1} & \review{48.6 $\pm$ 0.4} & \review{41.2 $\pm$ 0.0} & \review{64.8}\\
            mDSDI~\cite{msdi} & I & 79.0 $\pm$ 0.3 & 86.2 $\pm$ 0.2 & 69.2 $\pm$ 0.4 & 48.1 $\pm$ 1.4 & 42.8 $\pm$ 0.1 & 65.1\\
            GVRT (PTE)~\cite{gvrt} & I+T & 79.0 $\pm$ 0.2 & 85.1 $\pm$ 0.3 & 70.1 $\pm$ 0.1 & 48.0 $\pm$ 0.2 & 44.1 $\pm$ 0.1 & 65.2\\
            MIRO~\cite{miro} & I & 79.0 $\pm$ 0.0 & 85.4 $\pm$ 0.4 & 70.5 $\pm$ 0.4 & 50.4 $\pm$ 1.1 & 44.3 $\pm$ 0.2 & \textbf{65.9}\\
            \midrule            
            Ours$^{\dagger}$ & I+T & 78.3 $\pm$ 0.4 & 85.7 $\pm$ 0.1 & 70.1 $\pm$ 0.1 & 49.5 $\pm$ 0.9 & 43.7 $\pm$ 0.0 & 65.5 \\
            Ours$^{\ddagger}$ & I+T & 78.6 $\pm$ 0.3 & 87.0 $\pm$ 0.4 & 70.4 $\pm$ 0.2 & 49.2 $\pm$ 0.5 & 44.2 $\pm$ 0.0 & \textbf{65.9}\\
            \bottomrule
            \end{tabular}}
        
        \end{center}
        \vspace{-2.5em}
    \end{table}
    
}
\vspace{-1.0em}
\subsubsection{Ablation Studies.} 
In Table~\ref{tab:ablation-module}, we conduct an ablation study to demonstrate the effect of main modules: (i) a global alignment, (ii) a local alignment, (iii) a visual graph, and (iv) a textual graph. Our study demonstrates that (1) a global alignment, which aligns graph-level features together, effectively improves accuracies, especially in photo, cartoon, and art domains. (2) Adding local alignment, which aligns graphs via the clustering-based matching algorithm, improves all domains while using both alignments outperforms the alternatives. (3) Either using a visual or textual graph alone improves model generalization, but the gain is marginal with the visual graph alone. (4) The gain is maximized by using both graphs, which indicates that a graph structure effectively transfers text knowledge to train a generalizable visual encoder.

\setlength{\tabcolsep}{11pt}
\renewcommand{\arraystretch}{1.1} 
\begin{table}[t]
    \centering
    \caption{\review{Performance comparison between variants of our model with different matching techniques: bipartite matching and greedy matching. Data: CUB-DG.}}
    \label{tab:ablation-matching}
    \resizebox{\linewidth}{!}{%
    \begin{tabular}{cccccccc}
        \toprule
        \multirow{2}{*}{\shortstack{\review{Bipartite}\\\review{Matching}}} &
        \multirow{2}{*}{\shortstack{\review{Greedy}\\\review{Matching}}} &
        \multirow{2}{*}{\shortstack{\review{Time}\\\review{Complexity}}} &
        \multicolumn{4}{c}{\review{Target Domain}} &
        \multirow{2}{*}{\review{Avg.}} \\
        \cmidrule{4-7}
        & & & \review{Photo} & \review{Cartoon} & \review{Art} & \review{Paint} &  \\\midrule
        \review{-} & \review{\checkmark} & \review{$O(V^2)$} & \review{75.3} & \review{64.7} & \review{\textbf{54.8}} & \review{36.6} & \review{57.8} \\
        \review{\checkmark} & \review{-} & \review{$O(V^3)$} & \review{\textbf{75.4}} & \review{\textbf{65.5}} & \review{54.0} & \review{\textbf{41.4}} & \review{\textbf{59.1}} \\
        \bottomrule
        \vspace{-3em}
    \end{tabular}}
\end{table} 

\subsubsection{\review{Complexity of Graph Matching.}}
\review{
In our matching algorithm design, we explored two approaches: bipartite matching and greedy matching. Bipartite matching establishes one-to-one correspondences between clusters, minimizing pairwise distances, while greedy matching allows many-to-one associations based on spatial proximity.
Bipartite matching operates with a time complexity of $O(V^3)$, where $V$ denotes the number of vertices, while greedy matching operates in $O(V^2)$.
Despite the higher time complexity of bipartite matching, our experimental results (refer to Table~\ref{tab:ablation-matching}) demonstrate its superior performance over greedy matching.
This may be attributed to the constraints imposed on one-to-one matching, which result in a dispersed effect on models attempting to optimize cluster pairs on a global scale. 
Moreover, given that $V$ does not exceed 5 in our method, we opted for bipartite matching due to its enhanced performance in our specific context.
}
%
\subsection{Performance on DomainBed Benchmark}
We evaluate our model with a large-scale DomainBed~\cite{gulrajani2020search} datasets. We use the following five multi-domain datasets, including VLCS~\cite{fang2013unbiased}, PACS~\cite{Li2017dg}, OfficeHome~\cite{venkateswara2017deep}, TerraIncognita~\cite{beery2018recognition}, and DomainNet~\cite{peng2019moment}, comparing ours with 19 domain generalization algorithms. Due to space constraints, we only report \review{top-6} results (see supplementary material for full table).
The reported score represents averaged results obtained from three independent runs using randomly chosen hyperparameters.
We observe in Table~\ref{tab:performance_domainbed} that our proposed method shows matched or better state-of-the-art performance, where it ranks 1st (tied) in average performance.

%
\vspace{-0.5em}
\section{Conclusion}
We propose a novel domain generalization method that encodes domain-invariant visual representations.
To this end, we use a textual description to utilize verbalized (domain-invariant) knowledge from humans' typical reasoning. To align these, we use a clustering-based graph-matching algorithm based on visual and textual graphs built upon images and texts, respectively. We evaluate our model with state-of-the-art domain generalization approaches on CUB-DG and DomainBed datasets, achieving SOTA performance.

\vspace{-0.5em}
\subsubsection{Acknowledgements} 
This work was supported by Basic Science Research Program through the National Research Foundation of Korea(NRF) funded by the Ministry of Education(NRF-2021R1A6A1A13044830, 20\%) and supported by Institute of Information \& communications Technology Planning \& Evaluation(IITP) grant funded by the Korea government(MSIT) (RS-2022-II220043, Adaptive Personality for Intelligent Agents, 30\%, IITP-2024-RS-2024-00397085, Leading Generative AI Human Resources Development, 30\%, RS-2202-II220264, Comprehensive Video Understanding and Generation with Knowledge-based Deep Logic Neural Network, 20\%). Daewon Chae was supported by Hyundai Motor Chung Mong-Koo Foundation. 




%
\title{Clustering-based Image-Text Graph Matching for Domain Generalization}

\authorrunning{N. Park et al.}
\author{}
\institute{}
\maketitle 

\vspace{-1.0em}
In this supplementary material, we offer additional details that couldn't be accommodated in the main manuscript due to space limitations.
Specifically, we provide a comprehensive exploration of the graph-based visual representation (Section~\ref{sec:visualgraph}), implementation specifics (Section~\ref{sec:implementdetails}), examples of descriptions generated using InstructBLIP (Section~\ref{sec:instructblip}), the full performance table for CUB-DG (Section~\ref{sec:cubdg_full}), detailed t-SNE visualizations on CUB-DG (Section~\ref{sec:tsne}), analysis of the number of attributes in text data (Section~\ref{sec:diverse_attribute}), detailed DomainBed experiment results (Section~\ref{sec:domainbed_full}), and analysis of the PACS dataset (Section~\ref{sec:sup_pacs}).

\section{Graph-based Visual Representation}
\label{sec:visualgraph}
\subsubsection{Locally-aware Visual Graph Construction.}
\begin{figure}[H]
    \centering
    \begin{subfigure}{\columnwidth}
        \centering
        \includegraphics[width=.6\linewidth]{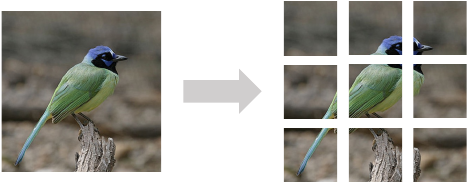}
        \caption{Split the image into $M$ grids.}
        \label{fig:grids}
    \end{subfigure}
    \begin{subfigure}{\columnwidth}
        \centering
        \vspace{1.5em}
        \includegraphics[width=.6\linewidth]{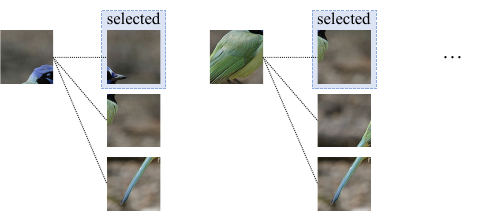}
        \caption{Calculate the $L_2$ distance between each grid and select $K_v$ nearest nodes. In this figure, $K_v$ is set to one.}
        \label{fig:l2}
    \end{subfigure}
    \begin{subfigure}{\columnwidth}
        \centering
        \vspace{1.5em}
        \includegraphics[width=.6\linewidth]{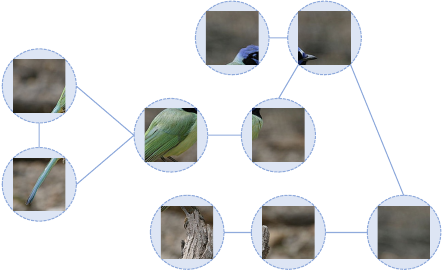}
        \caption{Construct the graph based on (a) and (b)}
        \label{fig:visualgraph}
    \end{subfigure}
    \caption{3 Steps to construct Locally-aware Visual Graph}
    \label{fig:3_steps}
    \vspace{-2.5em}
\end{figure}

In this section, we describe more detailed process of constructing the locally-aware visual graph.
Upon inputting an image of dimension $\mathbb{R}^{H \times W \times 3}$ into the backbone (utilizing ResNet50~\cite{resnet50} pre-trained on the ImageNet~\cite{deng2009imagenet} dataset), a feature of dimension $\mathbb{R}^{m \times m \times d}$ is generated, where $m \times m = M$ (see Figure~\ref{fig:grids}). This feature is obtained prior to passing through the global average pooling layer of the backbone. Subsequently, this vector enters the global average pooling layer, resulting in a vector of dimension $\mathbb{R}^{d}$. At this stage, the $d$-dimensional vector becomes the global visual representation, while the $m \times m \times d$ dimensional vector is employed to construct the visual graph. Each grid corresponds to a node in the visual graph, possessing a $d$-dimensional feature. In our experiments, we set $m$ and $M$ to 14 and 196, respectively.
For each local visual representation, we compute the $L_2$ distance between itself and the remaining $M$-1 local visual representations. Subsequently, we consider the $K_v$ closest neighbors as adjacent nodes. This results in a total of $M \times K_v$ edges. Figure~\ref{fig:l2} illustrates the process of ranking nodes based on their $L_2$ distance from each node, with only the top one node selected (for explanatory purposes, we set $K_v$ to one in Figure~\ref{fig:l2}).
Ultimately, the locally-aware visual graph is constructed, comprising $M$ nodes with $K_v$ neighboring nodes (refer to Figure~\ref{fig:visualgraph}).

\subsubsection{Graph-based Visual Representation.}
As described in the paper, we introduced an additional classifier that takes $\textbf{g}_v$ as an input to effectively capture the class-discriminative features.
This classifier is a linear layer trained by the standard cross-entropy loss.
Analysis of the results presented in Table \ref{tab:graph_loss} shows the inferior performance when the aforementioned classifier is not trained, demonstrating that the classifier is crucial to performance.

\setlength{\tabcolsep}{13pt}
\renewcommand{\arraystretch}{1.3}
\begin{table}[b]
    \caption{Out-of-distribution test accuracies (in \%) on the CUB-DG dataset. We compare our model with and without graph-based visual representation classification.}
    \centering
    \resizebox{\linewidth}{!}{%
    \begin{tabular}{cccccc}
        \toprule
        \multirow{2}{*}{\shortstack{Classification \\ for $\textbf{g}_v$}} &
        \multicolumn{4}{c}{Target Domain} &
        \multirow{2}{*}{Avg.} \\
        \cmidrule{2-5}
        & Photo & Cartoon & Art & Paint &  \\ 
        \midrule
        - & 74.7 & 62.3 & 52.3 & 35.7 & 56.2 \\
        \checkmark & \textbf{75.4} & \textbf{65.5} & \textbf{54.0} & \textbf{41.4} & \textbf{59.1} \\
        \bottomrule
    \end{tabular}
    }
    
    \label{tab:graph_loss}
    \vspace{-1.5em}
\end{table}


\section{Implementation Details} 
\label{sec:implementdetails}

For the CUB-DG dataset, we configured the batch size to be 32 for each source domain, and set $m$ and $M$—representing the number of local visual representations—to 14 and 196, respectively. Specifically, we assigned values of 8 and 3 to $K_v$ and $K_t$ (representing the number of nearest neighbors for visual and textual representation, respectively), while setting $N_v$ and $N_t$ to 5 and 3, respectively.
In the context of the DomainBed benchmark, we adjusted $m$ and $M$ to 7 and 49, respectively. Other hyperparameters were determined by the seed provided in the DomainBed benchmark. We report the averaged results from three independent runs.


\begin{figure}[t]
    \begin{center}
        \includegraphics[width=1.0\linewidth]{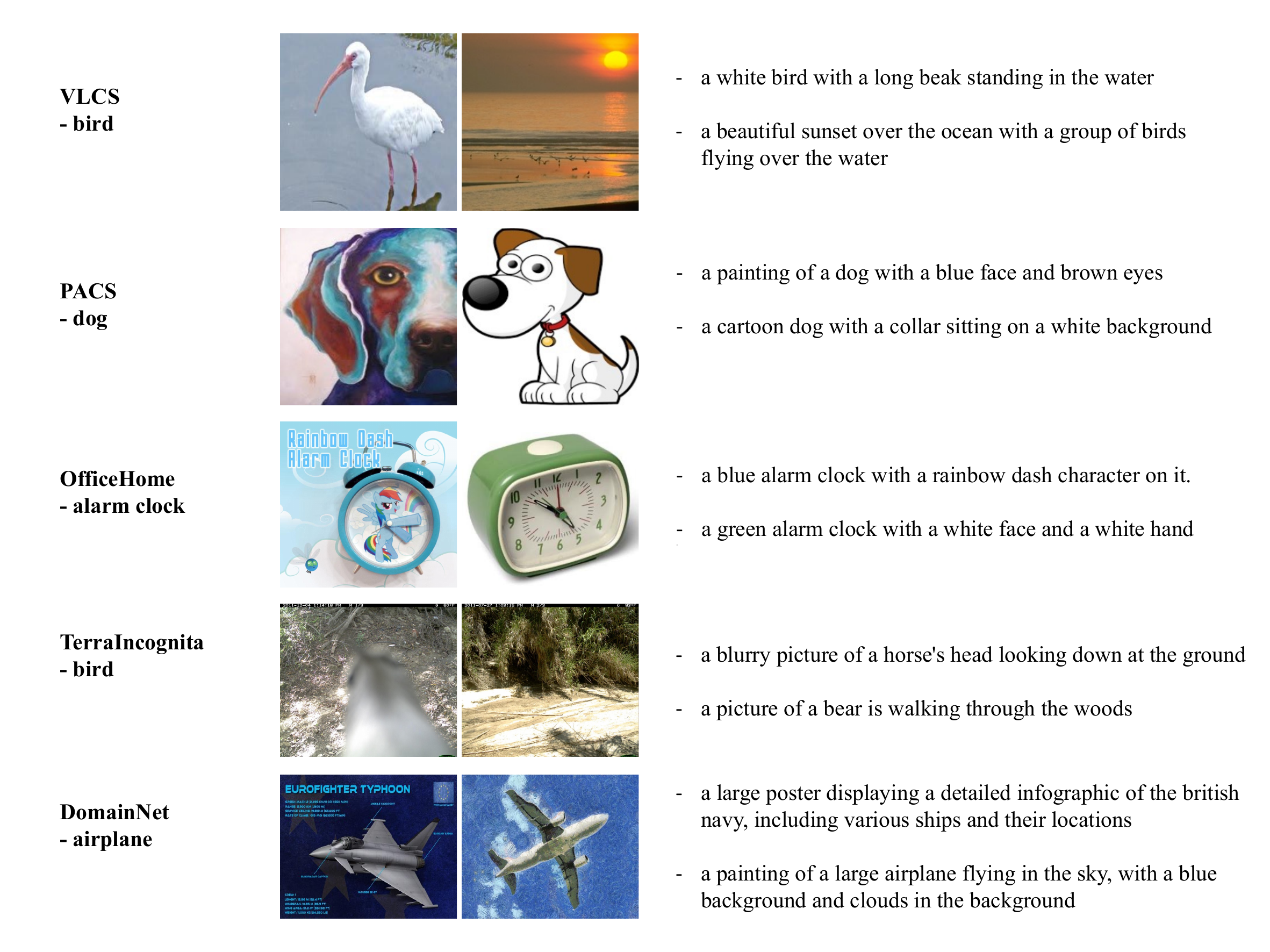}
    \end{center}
    \vspace{-2em}
    \caption{Example of a description generated by InstructBLIP for a randomly selected class and corresponding image for each dataset.}
    \label{fig:generated_description}
\end{figure}

\section{Descriptions Generated by InstructBLIP}
\label{sec:instructblip}

The datasets within the DomainBed benchmark do not include text inputs. To overcome this constraint, we have expanded our approach beyond class dictionary definitions by incorporating generated captions. To produce these descriptions, we leveraged InstructBLIP~\cite{instructblip}, setting a maximum token limit of 30 to ensure concise and relevant textual information.
Figure~\ref{fig:generated_description} shows examples of generated descriptions by InstructBLIP.

{
\setlength{\tabcolsep}{5pt}
\renewcommand{\arraystretch}{1.1} 
    \begin{table}[H]
    \vspace{-1.0em} 
    \caption{Full table for the out-of-distribution classification accuracies (in \%) on CUB-DG dataset. \textit{Abbr.} I: Image, T: Text.}
        \begin{center}
        \resizebox{.9\linewidth}{!}{
        \begin{tabular}{@{}lcccccc@{}}    \toprule
            \multirow{2}{*}{Model} & \multirow{2}{*}{Modality} &\multicolumn{4}{c}{Target Domain} & \multirow{2}{*}{Avg.} \\\cmidrule{3-6}
            & & Photo & Cartoon & Art & Paint & \\\midrule
            IRM~\cite{irm} & I & 60.6 & 51.6 & 36.5 & 30.3 & 44.8 \\
            GroupDRO~\cite{groupdro} & I & 60.9 & 54.8 & 36.5 & 27.0 & 44.8 \\
            ARM~\cite{arm} & I & 62.3 & 51.2 & 38.2 & 28.4 & 45.0 \\
            ERM~\cite{erm} & I & 62.5 & 53.2 & 37.4 & 29.0 & 45.5 \\
            VREx~\cite{vrex} & I & 63.9 & 54.9 & 38.6 & 30.1 & 46.9 \\
            CDANN~\cite{cdann} & I & 65.3 & 55.2 & 43.2 & 30.5 & 48.6 \\
            DANN~\cite{dann} & I & 67.5 & 57.0 & 42.8 & 30.6 & 49.5 \\
            Mixup~\cite{mixup} & I & 67.1 & 55.9 & 51.1 & 27.2 & 50.3 \\
            MixStyle~\cite{mixstyle} & I & 59.0 & 56.7 & 50.3 & 35.8 & 50.4 \\
            SagNet~\cite{sagnet} & I & 67.4 & 60.7 & 44.0 & 34.2 & 51.6 \\
            MIRO~\cite{miro} & I & 68.2 & 59.1 & 46.5 & 38.2 & 53.0 \\
            SD~\cite{sd} & I & 71.3 & 62.2 & 50.8 & 34.8 & 54.7 \\
            CORAL~\cite{coral} & I & 72.2 & 63.5 & 50.3 & 35.8 & 55.4 \\
            CCFP~\cite{CCFP} & I & 70.0 & 61.5 & 52.1 & 40.4 & 56.0 \\
            GVRT~\cite{gvrt} & I+T & 74.6 & 64.2 & 52.2 & 37.0 & 57.0 \\
            Ours & I+T & \textbf{75.4} & \textbf{65.5} & \textbf{54.0} & \textbf{41.4} & \textbf{59.1} \\
            & & (0.8\%$\uparrow$) & (1.3\%$\uparrow$) & (1.8\%$\uparrow$) &
            (1.0\%$\uparrow$) & (2.1\%$\uparrow$) \\
            \bottomrule
            \end{tabular}
            }
        \end{center}
        
        \vspace{-2.0em}
        \label{tab:performance_cub_all}
    \end{table}
}
\section{Peformance Comparison on CUB-DG} 
\label{sec:cubdg_full}

In Table~\ref{tab:performance_cub_all}, we provide our experiment on the CUB-DG dataset and compare it with the following 14 existing domain generalization algorithms, including GVRT~\cite{gvrt}, CORAL~\cite{coral}, SD~\cite{sd}, SagNet~\cite{sagnet}, MixStyle~\cite{mixstyle}, Mixup~\cite{mixup}, DANN~\cite{dann}, CDANN~\cite{cdann}, VREx~\cite{vrex}, ERM~\cite{erm}, ARM~\cite{arm}, GroupDRO~\cite{groupdro}, IRM~\cite{irm} and MIRO~\cite{miro}.
The reported numbers for MIRO is the result of tuning the hyperparameters to suit the CUB-DG dataset, and the rest are all taken from GVRT.


\section{Detailed t-SNE Visualizations on CUB-DG}
\label{sec:tsne}

\begin{figure}[t]
    \begin{center}
        \includegraphics[width=0.95\linewidth]{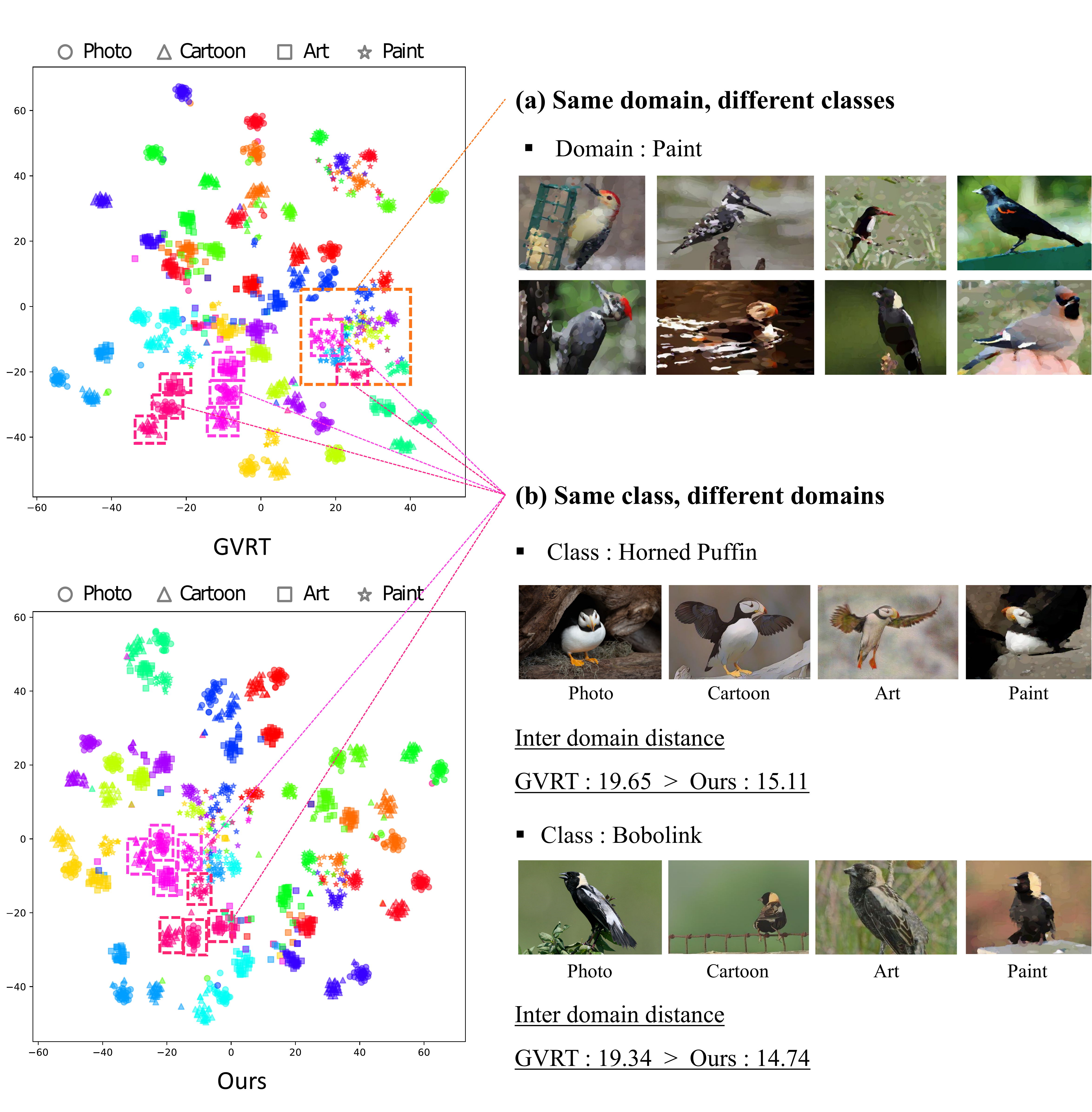}
    \end{center}
    \vspace{-1.5em}
    \caption{Visualization by t-SNE for GVRT and Ours with matched image samples.}
    \label{fig:tsne_supple}
    \vspace{-1.5em}
\end{figure}

In Figure \ref{fig:tsne_supple}, we provide a detailed t-SNE \cite{tsne} visualization of GVRT and Ours with matched image samples. Note that we mark different shapes to represent target domains and different colors to represent classes. In Figure \ref{fig:tsne_supple} (a), images that belong to the same domain (i.e. paint style) but different classes are gathered together in the GVRT feature space. Examining the corresponding images, they have their own class discriminative characteristics like the color of beak and pattern of feather, except that they share a common domain style. In other words, the features of images can be located far away if the class discriminative characteristic is captured. Therefore, it can be inferred that the GVRT model relies more on the domain-specific features rather than domain-invariant features for the images, limiting the ability of generalization. Figure \ref{fig:tsne_supple} (b) shows the distribution of images that belong to the same class but different domains. In our model, the features of same classes are located close each other unlike GVRT where the features of paint domain are located far away. In fact, our inter-domain distance is lower than GVRT. Thus, we can infer that ours captures more domain-invariant features than GVRT for the images. 



\begin{figure}[h]
    \begin{center}
        \includegraphics[width=0.95\linewidth]{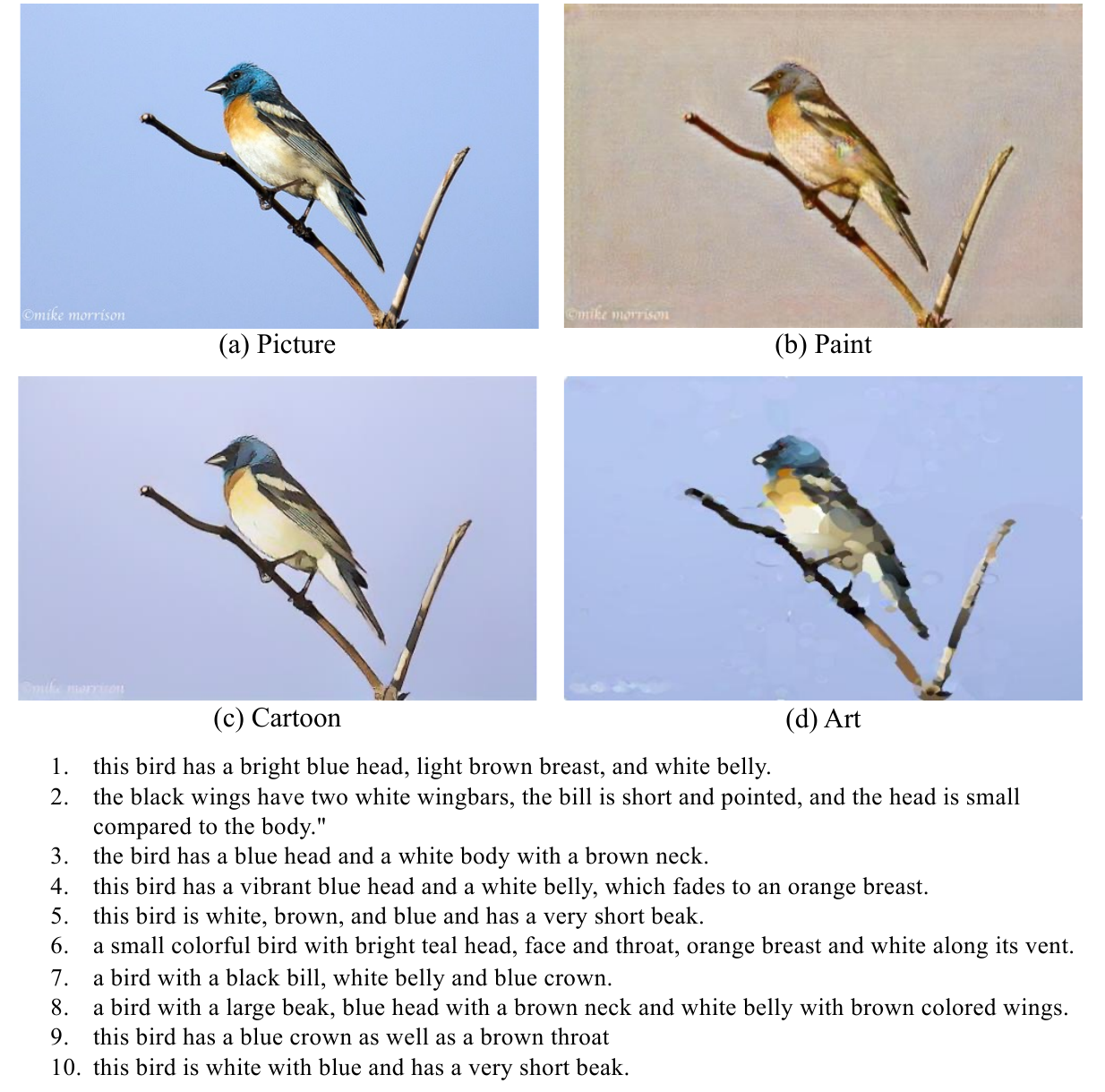}
    \end{center}
    \vspace{-2em}
    \caption{Example of CUB-DG Dataset.}
    \label{fig:cub_dg_sample}
    \vspace{-1em}
\end{figure}

\section{Analysis of the Number of Attributes in Text Data.}
\label{sec:diverse_attribute}
In CUB-DG dataset, each image has 10 text descriptions describing the content in detail. Each sentence contains more than one attribute. For example, in Figure~\ref{fig:cub_dg_sample}, first sentence has three attributes (e.g.,"bright blue head", "light brown breast" and "white belly"). To examine the impact of the number of attributes contained in text on model performance, we extracted a subset of data including 8018 instances containing sentences with 2, 3, and 4 attributes from the entire dataset. In Table~\ref{tab:diverse_attribute}, we observe that our proposed model improves performance when the number of attributes increases, but the base model~\cite{gvrt} does not.

{
\setlength{\tabcolsep}{10pt}
\renewcommand{\arraystretch}{1.1} 
\begin{table}[H]
    \vspace{-1em}
    \centering
    \caption{Performance comparison by number of attribute in text data on CUB-DG datasets.} 
    \vspace{1em}
    \label{tab:diverse_attribute}
    \resizebox{0.95\linewidth}{!}{
    \begin{tabular}{@{}lcccccc@{}}   
    \toprule
    \multirow{2}{*}{Algorithms} & \multirow{2}{*}{\shortstack{Attribute \\ Number}} &\multicolumn{4}{c}
    {Target Domain (Data: CUB-DG~\cite{gvrt})} & \multirow{2}{*}{Avg. $\uparrow$} \\\cmidrule{3-6}
    & & Photo & Cartoon & Art & Paint & \\\midrule
    & 2 & 71.26 & 59.94 & 47.11 & 32.82 & 52.78 \\
    GVRT~\cite{gvrt} & 3 & 71.59 & 59.64 & 46.42 & 33.35 & 52.75 \\
    & 4 & 71.85 & 59.93 & 45.74 & 32.85 & 52.59 \\
    \midrule
    & 2 & 72.56 & 62.27 & 48.38 & 35.75 & 54.74 \\
    Ours & 3 & \textbf{73.30} & 63.01 & 49.15 & 35.66 & 55.28 \\
    & 4 & 72.39 & \textbf{63.43} & \textbf{50.19} & \textbf{35.76} & \textbf{55.44} \\
    \bottomrule
    \end{tabular}
        }
\end{table}
}


\section{Performance on DomainBed Benchmark} 
\label{sec:domainbed_full}
Table~\ref{tab:performance_domainbed_full} shows the performance of our model and 19 different models in the DomainBed benchmark, and ours achieved state-of-the-art performance.
In Table \ref{tab:vlcs}–\ref{tab:domainnet}, we report per-domain results on each of the four multi-domain datasets from the large-scale DomainBed \cite{gulrajani2020search} experiments.
We provide the averaged results from three independent trials.
In each of the three trials, all choices, such as the dataset split, hyperparameter search, and weight initialization are selected randomly.
For model selection, we used the validation set from the source domains.
The reported numbers for SelfReg~\cite{selfreg}, and mDSI~\cite{msdi} were obtained from their respective papers, and the numbers for the remaining results were reported in the DomainBed~\cite{gulrajani2020search}. 
Note that GVRT and ours use multi-modal inputs (images and texts), while others only use images.


\begin{figure}[h]
    \centering
    \begin{subfigure}{.325\textwidth}
        \centering
        \includegraphics[width=\textwidth]{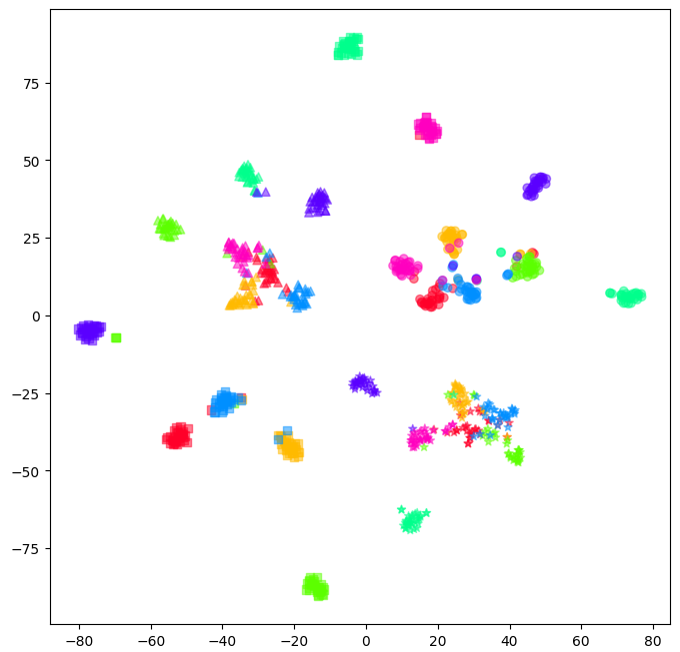}
        \caption{MIRO}
        \label{fig:miro_pacs_tsne}
    \end{subfigure}
    \begin{subfigure}{.325\textwidth}
        \centering
        \includegraphics[width=\textwidth]{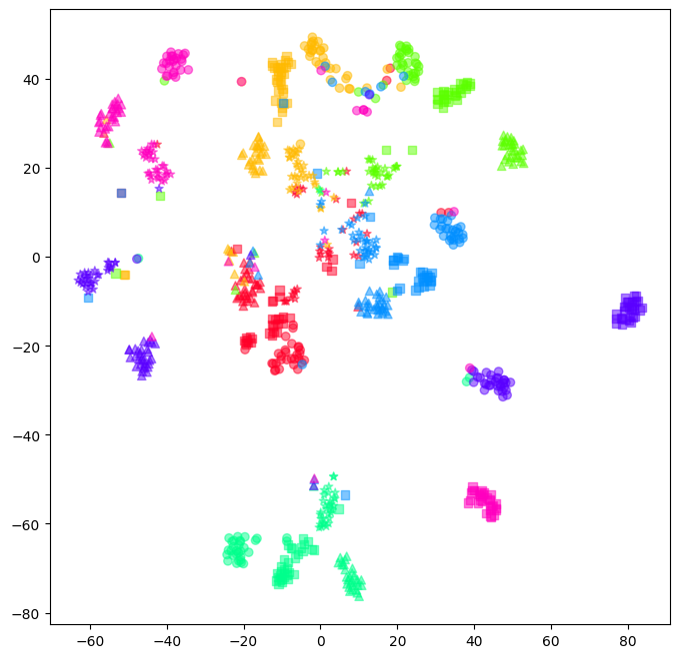}
        \caption{GVRT}
        \label{fig:gvrt_pacs_tsne}
    \end{subfigure}
    \begin{subfigure}{.325\textwidth}
        \centering
        \includegraphics[width=\textwidth]{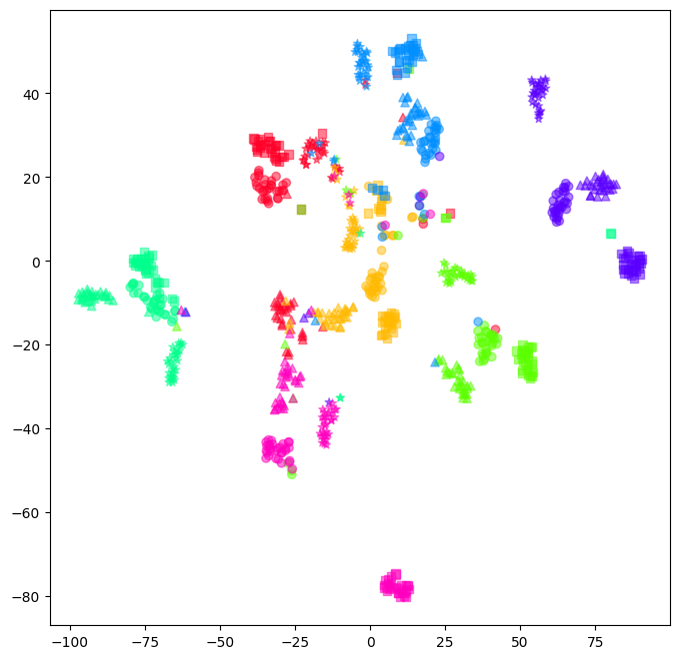}
        \caption{Ours}
        \label{fig:ours_pacs_tsne}
    \end{subfigure}
    \caption{Visualizations by t-SNE for (a) MIRO, (b) GVRT and (c) Ours.}
    \label{fig:pacs_tsne}
\end{figure}

\begin{figure}[h]
    \centering
    \includegraphics[width=\linewidth]{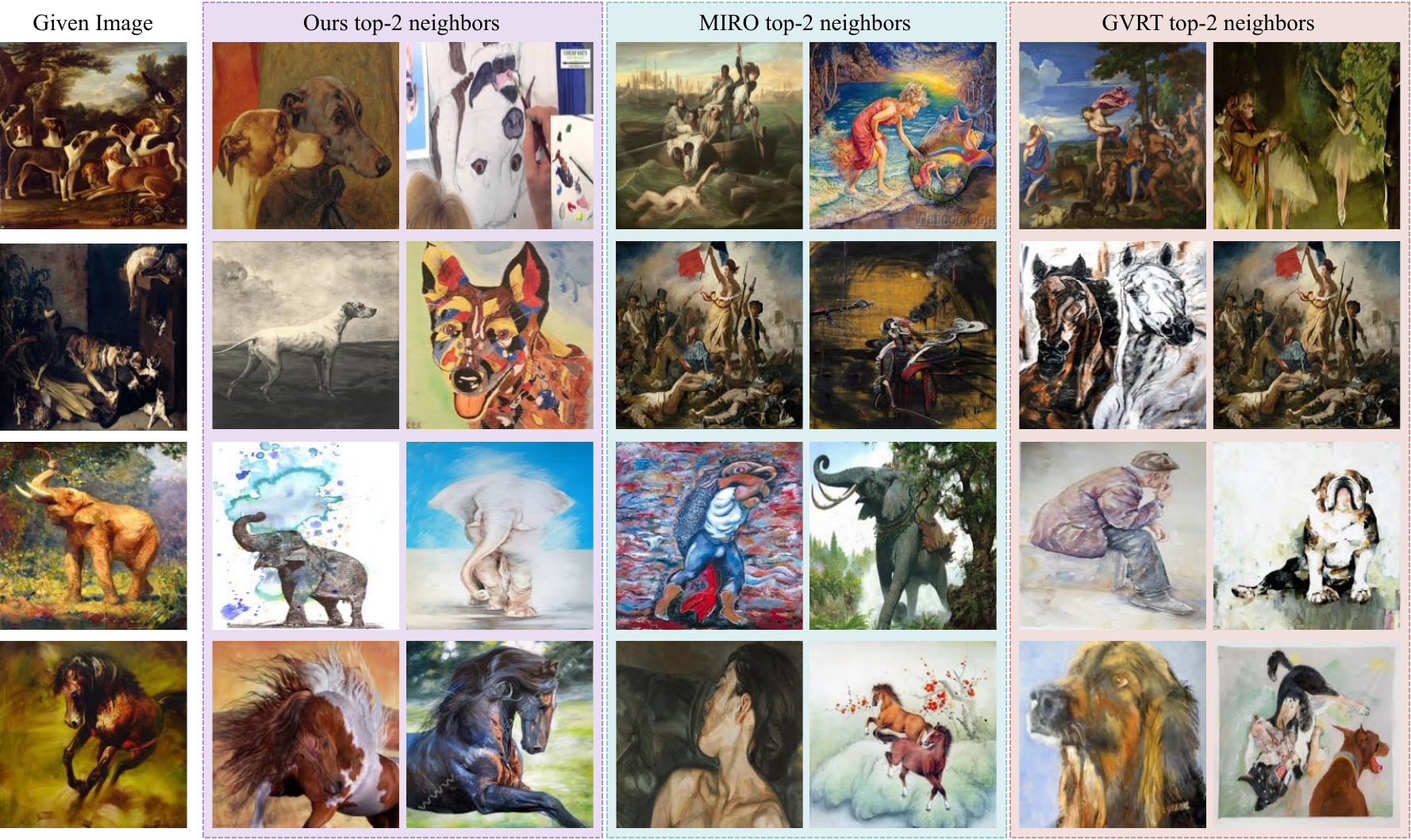}
    \caption{Exemplars of the top-2 nearest images to a specified image within the Art Painting domain of PACS dataset (e.g., ``dog'', ``dog'', ``elephant'', and ``horse'').}
    \label{fig:nearest_top2}
    \vspace{-.5em}
\end{figure}

\section{Analysis on PACS dataset}
\label{sec:sup_pacs}
In Figure~\ref{fig:pacs_tsne}, we present t-SNE visualizations for (a) MIRO, (b) GVRT, and (c) Ours to illustrate their embedding spaces. In these visualizations, we generated these visualizations using 30 data samples for each class within each domain, employing distinct marker styles to represent target domains and various colors to distinguish between classes. Notably, both GVRT and our model outperform MIRO. Furthermore, when comparing our model to GVRT, we observe distinct improvements. Specifically, in the case of purple points representing the `house' class and pink points corresponding to the `person' class in the GVRT visualization, they are noticeably scattered and distant from each other. In contrast, our model exhibits significantly improved clustering, leading to a more compact and coherent distribution of data points.
Figure~\ref{fig:nearest_top2} shows the top-2 nearest neighbors in the unseen target domain to the given images. As illustrated in the first row, our model demonstrates that the nearest neighbor images to the provided image all belong to the category of dogs, whereas MIRO and GVRT predominantly feature human images.


{
\setlength{\tabcolsep}{1.5pt}
\renewcommand{\arraystretch}{1.3} 
    \begin{table}[]
        \caption{Full table for the test accuracies (in \%) on the DomainBed benchmark in the multi-source DG task setting. \textit{Abbr.} I: Image, T: Text.}
        \label{tab:performance_domainbed_full}
        \begin{center}
        \resizebox{\linewidth}{!}{%
        \begin{tabular}{@{}lccccccc@{}}    \toprule
            \multirow{2}{*}{Algorithm} & \multirow{2}{*}{Modality} & \multicolumn{5}{c}{Dataset} & \multirow{2}{*}{Avg.} \\\cmidrule{3-7}
            & & VLCS & PACS & OfficeHome & TerraIncognita & DomainNet \\\midrule
            Ours (w/ InstructBLIP) & I+T & 78.6 $\pm$ 0.3 & 87.0 $\pm$ 0.4 & 70.4 $\pm$ 0.2 & 49.2 $\pm$ 0.5 & 44.2 $\pm$ 0.0 & \textbf{65.9} \\
            Ours (w/ dictionary) & I+T & 78.3 $\pm$ 0.4 & 85.7 $\pm$ 0.1 & 70.1 $\pm$ 0.1 & 49.5 $\pm$ 0.9 & 43.7 $\pm$ 0.0 & 65.5 \\
            \midrule
            GVRT (PTE)~\cite{gvrt} & I+T & 79.0 $\pm$ 0.2 & 85.1 $\pm$ 0.3 & 70.1 $\pm$ 0.1 & 48.0 $\pm$ 0.2 & 44.1 $\pm$ 0.1 & 65.2\\
            MIRO~\cite{miro} & I & 79.0 $\pm$ 0.0 & 85.4 $\pm$ 0.4 & 70.5 $\pm$ 0.4 & 50.4 $\pm$ 1.1 & 44.3 $\pm$ 0.2 & \textbf{65.9}\\
            mDSDI~\cite{msdi} & I & 79.0 $\pm$ 0.3 & 86.2 $\pm$ 0.2 & 69.2 $\pm$ 0.4 & 48.1 $\pm$ 1.4 & 42.8 $\pm$ 0.1 & 65.1\\
            CCFP~\cite{CCFP} & I & 78.9 $\pm$ 0.3 & 86.6 $\pm$ 0.2 & 68.9 $\pm$ 0.1 & 48.6 $\pm$ 0.4 & 41.2 $\pm$ 0.0 & 64.8 \\
            CORAL~\cite{coral} & I & 78.8 $\pm$ 0.6 & 86.2 $\pm$ 0.3 & 68.7 $\pm$ 0.3 & 47.6 $\pm$ 1.0 & 41.5 $\pm$ 0.1 & 64.6\\
            SagNet~\cite{sagnet} & I & 77.8 $\pm$ 0.5 & 86.3 $\pm$ 0.2 & 68.1 $\pm$ 0.1 & 48.6 $\pm$ 1.0 & 40.3 $\pm$ 0.1 & 64.2\\
            SelfReg~\cite{selfreg} & I & 77.8 $\pm$ 0.9 & 85.6 $\pm$ 0.4  & 67.9 $\pm$ 0.7 & 47.0 $\pm$ 0.3 & 42.8 $\pm$ 0.0 & 64.2\\
            Fish~\cite{fish} & I & 77.8 $\pm$ 0.3 & 85.5 $\pm$ 0.3 &  68.6 $\pm$ 0.4 & 45.1 $\pm$ 1.3 & 42.7 $\pm$ 0.2 & 63.9\\
            MLDG~\cite{mldg} & I & 77.2 $\pm$ 0.4 & 84.9 $\pm$ 1.0 & 66.8 $\pm$ 0.6 & 47.7 $\pm$ 0.9 & 41.2 $\pm$ 0.1 & 63.6\\
            Mixup~\cite{mixup} & I & 77.4 $\pm$ 0.6 & 84.6 $\pm$ 0.6 & 68.1 $\pm$ 0.3 & 47.9 $\pm$ 0.8 & 39.2 $\pm$ 0.1 & 63.4\\
            ERM~\cite{erm} & I & 77.5 $\pm$ 0.4 & 85.5 $\pm$ 0.2 & 66.5 $\pm$ 0.3 & 46.1 $\pm$ 1.8 & 40.9 $\pm$ 0.1 & 63.3\\
            MTL~\cite{mtl} & I & 77.2 $\pm$ 0.4 & 84.6 $\pm$ 0.5 & 66.4 $\pm$ 0.5 & 45.6 $\pm$ 1.2 & 40.6 $\pm$ 0.1 & 62.9\\
            RSC~\cite{rsc} & I & 77.1 $\pm$ 0.5 & 85.2 $\pm$ 0.9 & 65.5 $\pm$ 0.9 & 46.6 $\pm$ 1.0 & 38.9 $\pm$ 0.5 & 62.7\\
            DANN~\cite{dann} & I & 78.6 $\pm$ 0.4 & 83.6 $\pm$ 0.4 & 65.9 $\pm$ 0.6 & 46.7 $\pm$ 0.5 & 38.3 $\pm$ 0.1 & 62.6\\
            CDANN~\cite{cdann} & I & 77.5 $\pm$ 0.1 & 82.6 $\pm$ 0.9 & 65.8 $\pm$ 1.3 & 45.8 $\pm$ 1.6 & 38.3 $\pm$ 0.3 & 62.0 \\
            VREx~\cite{vrex} & I & 78.3 $\pm$ 0.2 & 84.9 $\pm$ 0.6 & 66.4 $\pm$ 0.6 & 46.4 $\pm$ 0.6 & 33.6 $\pm$ 2.9 & 61.9\\
            ARM~\cite{arm} & I & 77.6 $\pm$ 0.3 & 85.1 $\pm$ 0.4 & 64.8 $\pm$ 0.3 & 45.5 $\pm$ 0.3 & 35.5 $\pm$ 0.2 & 61.7\\
            IRM~\cite{irm} & I & 78.5 $\pm$ 0.5 & 83.5 $\pm$ 0.8 & 64.3 $\pm$ 2.2 & 47.6 $\pm$ 0.8 & 33.9 $\pm$ 2.8 & 61.6\\
            GroupDRO~\cite{groupdro} & I & 76.7 $\pm$ 0.6 & 84.4 $\pm$ 0.8 & 66.0 $\pm$ 0.7 & 43.2 $\pm$ 1.1 & 33.3 $\pm$ 0.2 & 60.7\\
            MMD~\cite{mmd} & I & 77.5 $\pm$ 0.9 & 84.6 $\pm$ 0.5 & 66.3 $\pm$ 0.1 & 42.2 $\pm$ 1.6 & 23.4 $\pm$ 9.5 & 58.8\\
            \bottomrule
            \end{tabular}}
        \end{center}
        
    \end{table}
}
{
\setlength{\tabcolsep}{5pt}
\renewcommand{\arraystretch}{1.2}
    \begin{table}[]
        \caption{
        Per-domain out-of-distribution test accuracies on the VLCS~\cite{fang2013unbiased} dataset. \textit{Abbr.} I: Image, T: Text}
        \label{tab:vlcs}
        \begin{center}
        \resizebox{\linewidth}{!}{%
        \begin{tabular}{lcccccc} \toprule
            Algorithm & Modality & Caltech & LabelMe & SUN09 & VOC2007 & Avg. \\\midrule
            Ours (w/ InstructBLIP) & I+T & 98.3 $\pm$ 0.1 & 64.5 $\pm$ 0.3 & 73.7 $\pm$ 0.8 & 77.8 $\pm$ 1.1 & 78.6 \\
            Ours (w/ dictionary) & I+T & 98.3 $\pm$ 0.3 & 64.6 $\pm$ 0.7 & 73.6 $\pm$ 2.2 & 76.6 $\pm$ 0.8 & 78.3 \\
            \midrule
            GVRT (PTE) ~\cite{gvrt} & I+T & 98.8 $\pm$ 0.1 & 64.0 $\pm$ 0.3 & 75.2 $\pm$ 0.5 & 77.9 $\pm$ 1.0 & 79.0\\
            MIRO~\cite{miro} & I & - & - & - & - & 79.0\\
            mDSDI~\cite{msdi} & I & 97.6 $\pm$ 0.1 & 66.5 $\pm$ 0.4 & 74.0 $\pm$ 0.6 & 77.8 $\pm$ 0.7 & 79.0\\
            CCFP~\cite{CCFP} & I & 98.1 $\pm$ 0.2 & 649 $\pm$ 0.1 & 78.3 $\pm$ 0.2 & 74.5 $\pm$ 1.5 & 78.9\\
            CORAL~\cite{coral} & I & 98.3 $\pm$ 0.1       & 66.1 $\pm$ 1.2       & 73.4 $\pm$ 0.3       & 77.5 $\pm$ 1.2       & 78.8\\
            DANN~\cite{dann} & I & 99.0 $\pm$ 0.3       & 65.1 $\pm$ 1.4       & 73.1 $\pm$ 0.3       & 77.2 $\pm$ 0.6       & 78.6\\
            IRM~\cite{irm} & I & 98.6 $\pm$ 0.1       & 64.9 $\pm$ 0.9       & 73.4 $\pm$ 0.6       & 77.3 $\pm$ 0.9       & 78.5\\
            VREx~\cite{vrex} & I & 98.4 $\pm$ 0.3       & 64.4 $\pm$ 1.4       & 74.1 $\pm$ 0.4       & 76.2 $\pm$ 1.3       & 78.3\\
            SelfReg~\cite{selfreg} & I & 96.7 $\pm$ 0.4 & 65.2 $\pm$ 1.2  & 73.1 $\pm$ 1.3 & 76.2 $\pm$ 0.7 & 77.8\\    
            SagNet~\cite{sagnet} & I & 97.9 $\pm$ 0.4       & 64.5 $\pm$ 0.5       & 71.4 $\pm$ 1.3       & 77.5 $\pm$ 0.5       & 77.8 \\
            Fish~\cite{fish} & I & - & - & - & - & 77.8\\  
            ARM~\cite{arm} & I & 98.7 $\pm$ 0.2       & 63.6 $\pm$ 0.7       & 71.3 $\pm$ 1.2       & 76.7 $\pm$ 0.6       & 77.6\\
            MMD~\cite{mmd} & I & 97.7 $\pm$ 0.1       & 64.0 $\pm$ 1.1       & 72.8 $\pm$ 0.2       & 75.3 $\pm$ 3.3       & 77.5\\
            CDANN~\cite{cdann} & I & 97.1 $\pm$ 0.3       & 65.1 $\pm$ 1.2       & 70.7 $\pm$ 0.8       & 77.1 $\pm$ 1.5       & 77.5\\
            ERM~\cite{erm} & I & 97.7 $\pm$ 0.4       & 64.3 $\pm$ 0.9       & 73.4 $\pm$ 0.5       & 74.6 $\pm$ 1.3       & 77.5 \\
            Mixup~\cite{mixup} & I & 98.3 $\pm$ 0.6       & 64.8 $\pm$ 1.0       & 72.1 $\pm$ 0.5       & 74.3 $\pm$ 0.8       & 77.4\\ 
            MTL~\cite{mtl} & I & 97.8 $\pm$ 0.4       & 64.3 $\pm$ 0.3       & 71.5 $\pm$ 0.7       & 75.3 $\pm$ 1.7       & 77.2\\
            MLDG~\cite{mldg} & I & 97.4 $\pm$ 0.2       & 65.2 $\pm$ 0.7       & 71.0 $\pm$ 1.4       & 75.3 $\pm$ 1.0       & 77.2\\
            RSC~\cite{rsc} & I & 97.9 $\pm$ 0.1       & 62.5 $\pm$ 0.7       & 72.3 $\pm$ 1.2       & 75.6 $\pm$ 0.8       & 77.1\\
            GroupDRO~\cite{groupdro} & I & 97.3 $\pm$ 0.3       & 63.4 $\pm$ 0.9       & 69.5 $\pm$ 0.8       & 76.7 $\pm$ 0.7       & 76.7\\
            \bottomrule
            \end{tabular}}
        \end{center}
        
    \end{table}
}
{
\setlength{\tabcolsep}{5pt}
\renewcommand{\arraystretch}{1.2}
    \begin{table}[]
        \caption{
        Per-domain out-of-distribution test accuracies on the PACS~\cite{Li2017dg} dataset. \textit{Abbr.} I: Image, T: Text}
        \label{tab:pacs}
        \begin{center}
        \resizebox{\linewidth}{!}{%
        \begin{tabular}{lcccccc} \toprule
            Algorithm & Modality & Art Painting & Cartoon & Photo & Sketch & Avg. \\\midrule
            Ours (w/ InstructBLIP) & I+T & 87.9 $\pm$ 0.7 & 81.4 $\pm$ 0.1 & 98.0 $\pm$ 0.1 & 80.5 $\pm$ 1.1 & 87.0 \\
            Ours (w/ dictionary) & I+T & 87.1 $\pm$ 0.5 & 79.8 $\pm$ 0.4 & 97.7 $\pm$ 0.1 & 78.3 $\pm$ 0.7 & 85.7 \\
            \midrule
            GVRT (PTE) ~\cite{gvrt} & I+T & 87.9 $\pm$ 0.3 & 78.4 $\pm$ 1.0 & 98.2 $\pm$ 0.1 & 75.7 $\pm$ 0.4 & 85.1 \\
            CCFP~\cite{CCFP} & I & 87.5 $\pm$ 0.1 & 81.3 $\pm$ 0.3 & 96.4 $\pm$ 0.3 & 81.4 $\pm$ 0.8 & 86.6\\
            SagNet~\cite{sagnet} & I & 87.4 $\pm$ 1.0       & 80.7 $\pm$ 0.6       & 97.1 $\pm$ 0.1       & 80.0 $\pm$ 0.4       & 86.3\\
            mDSDI~\cite{msdi} & I & 87.7 $\pm$ 0.4 & 80.4 $\pm$ 0.7 & 98.1 $\pm$ 0.3 & 78.4 $\pm$ 1.2 & 86.2\\
            CORAL~\cite{coral} & I & 88.3 $\pm$ 0.2       & 80.0 $\pm$ 0.5       & 97.5 $\pm$ 0.3       & 78.8 $\pm$ 1.3       & 86.2\\
            SelfReg~\cite{selfreg} & I & 87.9 $\pm$ 1.0 & 79.4 $\pm$ 1.4  & 96.8 $\pm$ 0.7 & 78.3 $\pm$ 1.2 & 85.6\\
            ERM~\cite{erm} & I & 84.7 $\pm$ 0.4       & 80.8 $\pm$ 0.6       & 97.2 $\pm$ 0.3       & 79.3 $\pm$ 1.0       & 85.5\\
            Fish~\cite{fish} & I & - & - & - & - & 85.5\\
            MIRO~\cite{miro} & I & - & - & - & - & 85.4\\
            RSC~\cite{rsc} & I & 85.4 $\pm$ 0.8       & 79.7 $\pm$ 1.8       & 97.6 $\pm$ 0.3       & 78.2 $\pm$ 1.2       & 85.2\\
            ARM~\cite{arm} & I & 86.8 $\pm$ 0.6       & 76.8 $\pm$ 0.5       & 97.4 $\pm$ 0.3       & 79.3 $\pm$ 1.2       & 85.1\\
            VREx~\cite{vrex} & I & 86.0 $\pm$ 1.6       & 79.1 $\pm$ 0.6       & 96.9 $\pm$ 0.5       & 77.7 $\pm$ 1.7       & 84.9\\
            MLDG~\cite{mldg} & I & 85.5 $\pm$ 1.4       & 80.1 $\pm$ 1.7       & 97.4 $\pm$ 0.3       & 76.6 $\pm$ 1.1       & 84.9\\
            MMD~\cite{mmd} & I & 86.1 $\pm$ 1.4       & 79.4 $\pm$ 0.9       & 96.6 $\pm$ 0.2       & 76.5 $\pm$ 0.5       & 84.6 \\
            MTL~\cite{mtl} & I & 87.5 $\pm$ 0.8       & 77.1 $\pm$ 0.5       & 96.4 $\pm$ 0.8       & 77.3 $\pm$ 1.8       & 84.6\\
            Mixup~\cite{mixup} & I & 86.1 $\pm$ 0.5       & 78.9 $\pm$ 0.8       & 97.6 $\pm$ 0.1       & 75.8 $\pm$ 1.8       & 84.6 \\ 
            GroupDRO~\cite{groupdro} & I & 83.5 $\pm$ 0.9       & 79.1 $\pm$ 0.6       & 96.7 $\pm$ 0.3       & 78.3 $\pm$ 2.0       & 84.4\\
            DANN~\cite{dann} & I & 86.4 $\pm$ 0.8       & 77.4 $\pm$ 0.8       & 97.3 $\pm$ 0.4       & 73.5 $\pm$ 2.3       & 83.6\\
            IRM~\cite{irm} & I & 84.8 $\pm$ 1.3       & 76.4 $\pm$ 1.1       & 96.7 $\pm$ 0.6       & 76.1 $\pm$ 1.0       & 83.5\\
            CDANN~\cite{cdann} & I & 84.6 $\pm$ 1.8       & 75.5 $\pm$ 0.9       & 96.8 $\pm$ 0.3       & 73.5 $\pm$ 0.6       & 82.6\\
            \bottomrule
            \end{tabular}}
        \end{center}
        
    \end{table}
}
{
\setlength{\tabcolsep}{5pt}
\renewcommand{\arraystretch}{1.2}
    \begin{table}[]
        \caption{
        Per-domain out-of-distribution test accuracies on the OfficeHome~\cite{venkateswara2017deep} dataset. \textit{Abbr.} I: Image, T: Text}
        \begin{center}
        \resizebox{\linewidth}{!}{%
        \begin{tabular}{lcccccc} \toprule
            Algorithm & Modality & Art & Clipart & Product & Real-world & Avg. \\\midrule
            Ours (w/ InstructBLIP) & I+T & 66.5 $\pm$ 0.4 & 56.4 $\pm$ 0.4 & 78.5 $\pm$ 0.5 & 80.1 $\pm$ 0.1 & 70.4 \\
            Ours (w/ dictionary) & I+T & 66.7 $\pm$ 1.0 & 55.4 $\pm$ 0.4 & 78.2 $\pm$ 0.4 & 80.0 $\pm$ 0.3 & 70.1 \\
            \midrule
            GVRT (PTE) ~\cite{gvrt} & I+T & 66.3 $\pm$ 0.1 & 55.8 $\pm$ 0.4 & 78.2 $\pm$ 0.4 & 80.4 $\pm$ 0.2 & 70.1 \\
            MIRO~\cite{miro} & I & - & - & - & - & 70.5 \\
            mDSDI~\cite{msdi} & I & 68.1 $\pm$ 0.3 & 52.1 $\pm$ 0.4 & 76.0 $\pm$ 0.2 & 80.4 $\pm$ 0.2 & 69.2\\
            CCFP~\cite{CCFP} & I & 63.7 $\pm$ 0.3 & 55.5 $\pm$ 0.3 & 77.2 $\pm$ 0.4 & 79.2 $\pm$ 0.3 & 68.9\\
            CORAL~\cite{coral} & I & 65.3 $\pm$ 0.4       & 54.4 $\pm$ 0.5       & 76.5 $\pm$ 0.1       & 78.4 $\pm$ 0.5       & 68.7\\
            Fish~\cite{fish} & I & - & - & - & - & 68.6\\
            Mixup~\cite{mixup} & I & 62.4 $\pm$ 0.8       & 54.8 $\pm$ 0.6       & 76.9 $\pm$ 0.3       & 78.3 $\pm$ 0.2       & 68.1\\ 
            SagNet~\cite{sagnet} & I & 63.4 $\pm$ 0.2       & 54.8 $\pm$ 0.4       & 75.8 $\pm$ 0.4       & 78.3 $\pm$ 0.3       & 68.1\\
            SelfReg~\cite{selfreg} & I & 63.6 $\pm$ 1.4 & 53.1 $\pm$ 1.0  & 76.9 $\pm$ 0.4 & 78.1 $\pm$ 0.4 & 67.9\\
            MLDG~\cite{mldg} & I & 61.5 $\pm$ 0.9       & 53.2 $\pm$ 0.6       & 75.0 $\pm$ 1.2       & 77.5 $\pm$ 0.4       & 66.8\\
            ERM~\cite{erm} & I & 61.3 $\pm$ 0.7       & 52.4 $\pm$ 0.3       & 75.8 $\pm$ 0.1       & 76.6 $\pm$ 0.3       & 66.5\\
            MTL~\cite{mtl} & I & 61.5 $\pm$ 0.7       & 52.4 $\pm$ 0.6       & 74.9 $\pm$ 0.4       & 76.8 $\pm$ 0.4       & 66.4\\
            VREx~\cite{vrex} & I & 60.7 $\pm$ 0.9       & 53.0 $\pm$ 0.9       & 75.3 $\pm$ 0.1       & 76.6 $\pm$ 0.5       & 66.4\\
            MMD~\cite{mmd} & I & 60.4 $\pm$ 0.2       & 53.3 $\pm$ 0.3       & 74.3 $\pm$ 0.1       & 77.4 $\pm$ 0.6       & 66.3\\
            GroupDRO~\cite{groupdro} & I & 60.4 $\pm$ 0.7       & 52.7 $\pm$ 1.0       & 75.0 $\pm$ 0.7       & 76.0 $\pm$ 0.7       & 66.0\\
            DANN~\cite{dann} & I & 59.9 $\pm$ 1.3       & 53.0 $\pm$ 0.3       & 73.6 $\pm$ 0.7       & 76.9 $\pm$ 0.5       & 65.9\\
            CDANN~\cite{cdann} & I & 61.5 $\pm$ 1.4       & 50.4 $\pm$ 2.4       & 74.4 $\pm$ 0.9       & 76.6 $\pm$ 0.8       & 65.8\\
            RSC~\cite{rsc} & I & 60.7 $\pm$ 1.4       & 51.4 $\pm$ 0.3       & 74.8 $\pm$ 1.1       & 75.1 $\pm$ 1.3       & 65.5 \\
            ARM~\cite{arm} & I & 58.9 $\pm$ 0.8       & 51.0 $\pm$ 0.5       & 74.1 $\pm$ 0.1       & 75.2 $\pm$ 0.3       & 64.8\\
            IRM~\cite{irm} & I & 58.9 $\pm$ 2.3       & 52.2 $\pm$ 1.6       & 72.1 $\pm$ 2.9       & 74.0 $\pm$ 2.5       & 64.3\\
            \bottomrule
            \end{tabular}}
        \end{center}
        
        \label{tab:officehome}
    \end{table}
}
{
\setlength{\tabcolsep}{5pt}
\renewcommand{\arraystretch}{1.2}
    \begin{table}[]
        \caption{
        Per-domain out-of-distribution test accuracies on the TerraIncognita~\cite{beery2018recognition} dataset. \textit{Abbr.} I: Image, T: Text}
        \label{tab:terraincognita}
        \begin{center}
        \resizebox{\linewidth}{!}{%
        \begin{tabular}{lcccccc} \toprule
            Algorithm & Modality & L100 & L38 & L43 & L46 & Avg. \\\midrule
            Ours (w/ InstructBLIP) & I+T & 54.5 $\pm$ 0.6 & 46.7 $\pm$ 0.8 & 57.1 $\pm$ 1.2	& 39.0 $\pm$ 0.8 & 49.2 \\
            Ours (w/ dictionary) & I+T & 56.9 $\pm$ 3.0 & 45.5 $\pm$ 0.7 & 57.7 $\pm$ 1.4 & 37.8 $\pm$ 0.8 & 49.5 \\
            \midrule
            GVRT (PTE) ~\cite{gvrt} & I+T & 53.9 $\pm$ 1.3 & 41.8 $\pm$ 1.2 & 58.2 $\pm$ 0.9 & 38.0 $\pm$ 0.6 & 48.0 \\
            MIRO~\cite{miro} & I & - & - & - & - & 50.4\\
            SagNet~\cite{sagnet} & I & 53.0 $\pm$ 2.9       & 43.0 $\pm$ 2.5       & 57.9 $\pm$ 0.6       & 40.4 $\pm$ 1.3       & 48.6\\
            CCFP~\cite{CCFP} & I & 56.4 $\pm$ 1.8 & 42.3 $\pm$ 0.1 & 58.0 $\pm$ 0.7 & 37.5 $\pm$ 0.4 & 48.6\\
            mDSDI~\cite{msdi} & I & 53.2 $\pm$ 3.0 & 43.3 $\pm$ 1.0 & 56.7 $\pm$ 0.5 & 39.2 $\pm$ 1.3 & 48.1\\
            Mixup~\cite{mixup} & I & 59.6 $\pm$ 2.0       & 42.2 $\pm$ 1.4       & 55.9 $\pm$ 0.8       & 33.9 $\pm$ 1.4       & 47.9\\ 
            MLDG~\cite{mldg} & I & 54.2 $\pm$ 3.0       & 44.3 $\pm$ 1.1       & 55.6 $\pm$ 0.3       & 36.9 $\pm$ 2.2       & 47.7\\
            IRM~\cite{irm} & I & 54.6 $\pm$ 1.3       & 39.8 $\pm$ 1.9       & 56.2 $\pm$ 1.8       & 39.6 $\pm$ 0.8       & 47.6 \\
            CORAL~\cite{coral} & I & 51.6 $\pm$ 2.4       & 42.2 $\pm$ 1.0       & 57.0 $\pm$ 1.0       & 39.8 $\pm$ 2.9       & 47.6\\
            SelfReg~\cite{selfreg} & I & 48.8 $\pm$ 0.9 & 41.3 $\pm$ 1.8 & 57.3 $\pm$ 0.7 & 40.6 $\pm$ 0.9 & 47.0 \\
            DANN~\cite{dann} & I & 51.1 $\pm$ 3.5       & 40.6 $\pm$ 0.6       & 57.4 $\pm$ 0.5       & 37.7 $\pm$ 1.8       & 46.7\\
            RSC~\cite{rsc} & I & 50.2 $\pm$ 2.2       & 39.2 $\pm$ 1.4       & 56.3 $\pm$ 1.4       & 40.8 $\pm$ 0.6       & 46.6 \\
            VREx~\cite{vrex} & I & 48.2 $\pm$ 4.3       & 41.7 $\pm$ 1.3       & 56.8 $\pm$ 0.8       & 38.7 $\pm$ 3.1       & 46.4\\
            ERM~\cite{erm} & I & 49.8 $\pm$ 4.4       & 42.1 $\pm$ 1.4       & 56.9 $\pm$ 1.8       & 35.7 $\pm$ 3.9       & 46.1\\
            CDANN~\cite{cdann} & I & 47.0 $\pm$ 1.9       & 41.3 $\pm$ 4.8       & 54.9 $\pm$ 1.7       & 39.8 $\pm$ 2.3       & 45.8\\
            MTL~\cite{mtl} & I & 49.3 $\pm$ 1.2       & 39.6 $\pm$ 6.3       & 55.6 $\pm$ 1.1       & 37.8 $\pm$ 0.8       & 45.6\\
            ARM~\cite{arm} & I & 49.3 $\pm$ 0.7       & 38.3 $\pm$ 2.4       & 55.8 $\pm$ 0.8       & 38.7 $\pm$ 1.3       & 45.5\\
            Fish~\cite{fish} & I & - & - & - & - & 45.1\\
            GroupDRO~\cite{groupdro} & I & 41.2 $\pm$ 0.7       & 38.6 $\pm$ 2.1       & 56.7 $\pm$ 0.9       & 36.4 $\pm$ 2.1       & 43.2\\
            MMD~\cite{mmd} & I & 41.9 $\pm$ 3.0       & 34.8 $\pm$ 1.0       & 57.0 $\pm$ 1.9       & 35.2 $\pm$ 1.8       & 42.2\\
            \bottomrule
            \end{tabular}}
        \end{center}
        
    \end{table}
}
{
\setlength{\tabcolsep}{1pt}
\renewcommand{\arraystretch}{1.4}
    \begin{table}[]
        \caption{
        Per-domain out-of-distribution test accuracies on the DomainNet~\cite{peng2019moment} dataset. \textit{Abbr.} I: Image, T: Text}
        \label{tab:domainnet}
        \begin{center}
        \resizebox{\linewidth}{!}{%
        \begin{tabular}{lcccccccc} \toprule
            Algorithm & Modality & Clip & Info & Paint & Quick & Real & Sketch & Avg. \\\midrule
            Ours (w/ InstructBLIP) & I+T & 61.6 $\pm$ 0.2 & 21.1 $\pm$ 0.1 & 51.3 $\pm$ 0.1 & 13.9 $\pm$ 0.2 & 64.8 $\pm$ 0.1 & 52.5 $\pm$ 0.2 & 44.2 \\
            Ours (w/ dictionary) & I+T & 61.1 $\pm$ 0.1 & 20.4 $\pm$ 0.2 & 50.4 $\pm$ 0.1 & 13.5 $\pm$ 0.1 & 64.7 $\pm$ 0.3 & 51.9 $\pm$ 0.1 & 43.7 \\
            \midrule
            GVRT (PTE) ~\cite{gvrt} & I+T & 62.4 $\pm$ 0.4 & 21.0 $\pm$ 0.0 & 50.5 $\pm$ 0.4 & 13.8 $\pm$ 0.3 & 64.6 $\pm$ 0.4 & 52.4 $\pm$ 0.2 & 44.1 \\
            mDSDI~\cite{msdi} & I & 62.1 $\pm$ 0.3 & 19.1 $\pm$ 0.4 & 49.4 $\pm$ 0.4 & 12.8 $\pm$ 0.7 & 62.9 $\pm$ 0.3 & 50.4 $\pm$ 0.4 & 42.8\\
            CORAL~\cite{coral} & I & 59.2 $\pm$ 0.1 & 19.7 $\pm$ 0.2 & 46.6 $\pm$ 0.3 & 13.4 $\pm$ 0.4 & 59.8 $\pm$ 0.2 & 50.1 $\pm$ 0.6 & 41.5\\
            CCFP~\cite{CCFP} & I & 58.7 $\pm$ 0.2 & 19.4 $\pm$ 0.3 & 47.1 $\pm$ 0.3 & 13.4 $\pm$ 0.4 & 58.1 $\pm$ 0.4 & 50.5 $\pm$ 0.1 & 41.2\\
            SagNet~\cite{sagnet} & I & 57.7 $\pm$ 0.3 & 19.0 $\pm$ 0.2 & 45.3 $\pm$ 0.3 & 12.7 $\pm$ 0.5 & 58.1 $\pm$ 0.5 & 48.8 $\pm$ 0.2 & 40.3\\
            SelfReg~\cite{selfreg} & I & 60.7 $\pm$ 0.1 & 21.6 $\pm$ 0.1 & 49.4 $\pm$ 0.2 & 12.7 $\pm$ 0.1 & 60.7 $\pm$ 0.1 & 51.7 $\pm$ 0.1 & 42.8\\
            Mixup~\cite{mixup} & I & 55.7 $\pm$ 0.3 & 18.5 $\pm$ 0.5 & 44.3 $\pm$ 0.5 & 12.5 $\pm$ 0.4 & 55.8 $\pm$ 0.3 & 48.2 $\pm$ 0.5 & 39.2\\ 
            MLDG~\cite{mldg} & I & 59.1 $\pm$ 0.2 & 19.1 $\pm$ 0.3 & 45.8 $\pm$ 0.7 & 13.4 $\pm$ 0.3 & 59.6 $\pm$ 0.2 & 50.2 $\pm$ 0.4 & 41.2\\
            VREx~\cite{vrex} & I & 47.3 $\pm$ 3.5 & 16.0 $\pm$ 1.5 & 35.8 $\pm$ 4.6 & 10.9 $\pm$ 0.3 & 49.6 $\pm$ 4.9 & 42.0 $\pm$ 3.0 & 33.6\\
            ERM~\cite{erm} & I & 58.1 $\pm$ 0.3 & 18.8 $\pm$ 0.3 & 46.7 $\pm$ 0.3 & 12.2 $\pm$ 0.4 & 59.6 $\pm$ 0.1 & 49.8 $\pm$ 0.4 & 40.9\\
            DANN~\cite{dann} & I & 53.1 $\pm$ 0.2 & 18.3 $\pm$ 0.1 & 44.2 $\pm$ 0.7 &   11.8 $\pm$ 0.1 & 55.5 $\pm$ 0.4 & 46.8 $\pm$ 0.6 & 38.3\\
            RSC~\cite{rsc} & I & 55.0 $\pm$ 1.2 & 18.3 $\pm$ 0.5 & 44.4 $\pm$ 0.6 & 12.2 $\pm$ 0.2 & 55.7 $\pm$ 0.7 & 47.8 $\pm$ 0.9 & 38.9\\
            IRM~\cite{irm} & I & 48.5 $\pm$ 2.8 & 15.0 $\pm$ 1.5 & 38.3 $\pm$ 4.3 & 10.9 $\pm$ 0.5 & 48.2 $\pm$ 5.2 & 42.3 $\pm$ 3.1 & 33.9\\
            MTL~\cite{mtl} & I & 57.9 $\pm$ 0.5 & 18.5 $\pm$ 0.4 & 46.0 $\pm$ 0.1 & 12.5 $\pm$ 0.1 & 59.5 $\pm$ 0.3 & 49.2 $\pm$ 0.1 & 40.6\\
            ARM~\cite{arm} & I & 49.7 $\pm$ 0.3 & 16.3 $\pm$ 0.5 & 40.9 $\pm$ 1.1 & 9.4 $\pm$ 0.1 & 53.4 $\pm$ 0.4 & 43.5 $\pm$ 0.4 & 35.5\\
            CDANN~\cite{cdann} & I & 54.6 $\pm$ 0.4 & 17.3 $\pm$ 0.1 & 43.7 $\pm$ 0.9 & 12.1 $\pm$ 0.7 & 56.2 $\pm$ 0.4 & 45.9 $\pm$ 0.5 & 38.3\\
            MMD~\cite{mmd} & I & 32.1 $\pm$ 13.3 & 11.0 $\pm$ 4.6 & 26.8 $\pm$ 11.3 & 8.7 $\pm$ 2.1  & 32.7 $\pm$ 13.8 & 28.9 $\pm$ 11.9 & 23.4\\
            GroupDRO~\cite{groupdro} & I & 47.2 $\pm$ 0.5 & 17.5 $\pm$ 0.4 & 33.8 $\pm$ 0.5 & 9.3 $\pm$ 0.3 & 51.6 $\pm$ 0.4 & 40.1 $\pm$ 0.6 & 33.3\\
            MIRO~\cite{miro} & I & - & - & - & - & - & - & 44.3\\   
            Fish~\cite{fish} & I & - & - & - & - & - & - & 42.7\\  
            \bottomrule
            \end{tabular}}
        \end{center}
        
    \end{table}
}




{
    \small
    \bibliographystyle{splncs04}
    \bibliography{src/reference}
}
\end{document}